\documentclass{article}

\usepackage{microtype}
\usepackage{graphicx}
\usepackage{subcaption}
\usepackage{booktabs} 

\usepackage{hyperref}

\usepackage[preprint]{icml2026}

\usepackage{amsmath}
\usepackage{amssymb}
\usepackage{mathtools}
\usepackage{amsthm}
\usepackage{multicol}
\usepackage{multirow}
\usepackage[table]{xcolor}
\usepackage{diagbox}
\usepackage[capitalize,noabbrev]{cleveref}

\theoremstyle{plain}

\theoremstyle{definition}

\theoremstyle{remark}

\usepackage[textsize=tiny]{todonotes}

\icmltitlerunning{Evolution of Benchmark: Black-Box Optimization Benchmark Design through Large Language Model}

\begin{document}

\twocolumn[
  \icmltitle{Evolution of Benchmark: Black-Box Optimization Benchmark Design \\through Large Language Model}



  \icmlsetsymbol{equal}{*}

  \begin{icmlauthorlist}
    \icmlauthor{Chen Wang}{scut,equal}
    \icmlauthor{Sijie Ma}{scut,equal}
    \icmlauthor{Zeyuan Ma}{scut,equal}
    \icmlauthor{Yue-Jiao Gong}{scut}
  \end{icmlauthorlist}
  
  \icmlaffiliation{scut}{South China University of Technology}  

  \icmlcorrespondingauthor{Yue-Jiao Gong}{gongyuejiao@gmail.com}

  \icmlkeywords{Machine Learning, ICML}

  \vskip 0.3in
]



\printAffiliationsAndNotice{\icmlEqualContribution}

\begin{abstract}
Benchmark Design in Black-Box Optimization (BBO) is a fundamental yet open-ended topic. Early BBO benchmarks are predominantly human-crafted, introducing expert bias and constraining diversity. Automating this design process can relieve the human-in-the-loop burden while enhancing diversity and objectivity. We propose Evolution of Benchmark (EoB), an automated BBO benchmark designer empowered by the large language model (LLM) and its program evolution capability. Specifically, we formulate benchmark design as a bi-objective optimization problem towards maximizing (i) landscape diversity and (ii) algorithm-differentiation ability across a portfolio of BBO solvers. Under this paradigm, EoB iteratively prompts LLM to evolve a population of benchmark programs and employs  a reflection-based scheme to co-evolve the landscape and its corresponding program. Comprehensive experiments validate our EoB is a competitive candidate in multi-dimensional usages: 1) Benchmarking BBO algorithms; 2) Training and testing learning-assisted BBO algorithms; 3) Extending proxy for expensive real-world problems.
\end{abstract}

\section{Introduction}\label{sec:intro}
Black-Box Optimization~(BBO) problems are inherently hard~\cite{bbo-survey} and widely encountered across various real-world scenarios~\cite{engineer1,scientific1,es-llm}.
During the past two decades, a broad spectrum of BBO algorithms~\cite{1992ga,de,2002es,pso,bo,newton,optuna,vizier} have been proposed to address BBO problems~\cite{1992ga,de,2002es,pso,bo,newton}. 
However, compared to the exponential growth in algorithm development, the advancement of BBO benchmarks has lagged significantly behind.

A well-developed BBO benchmark is curcial for the advancement of the BBO field, as it helps ensure both fair comparison and performance coherence on out-of-benchmark tasks. Existing benchmarks such as CoCo-BBOB~\cite{bbob2010} and IEEE CEC series~\cite{cec2015} are commonly used within the community, where the synthetic functions are carefully designed to provide various landscape challenges~\cite{ela} for evaluating BBO algorithms. However, their limited testing instances hinder them from keeping pace with the swift emergence of novel BBO tasks in this fast-evolving era. Besides, given the rise of learning-assisted BBO techniques more recently~\cite{ma2025toward,metabbo-icl-5,GLHF}, training them requires a large number of diversified BBO instances to ensure robust generalization. This poses significant challenges for hand-crafted benchmark design since this paradigm needs deep expertise and massive labor input.

\begin{figure}[t]
\centering
\includegraphics[width=0.9\linewidth]{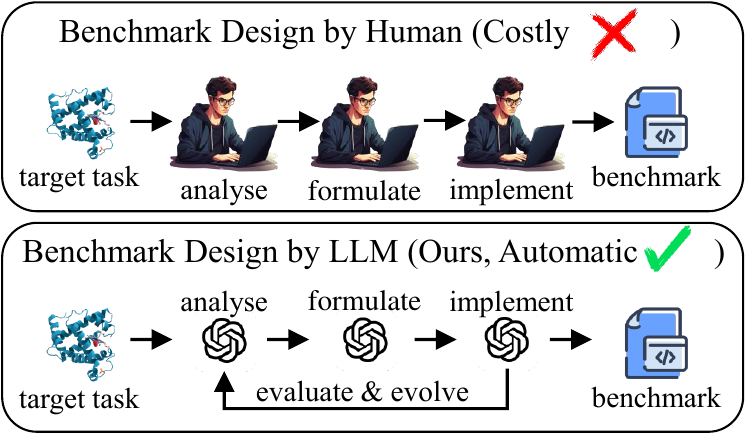}
\caption{Paradigm shift from human-crafted benchmark design toward LLM-based automation.}
\label{fig:intro}
\vspace{-3mm}
\end{figure}

To address these fundamental limitations, we propose Evolution of Benchmark~(EoB) as a compact, self-contained and fully automated BBO benchmark designer. As illustrated in Fig.~\ref{fig:intro}, EoB is empowered by recent advances of LLMs for program evolution~\cite{evoprompt,eoh,llamea,alphaevolve} to achieve both benchmark quality and  automated design efficiency. In such a paradigm, a population of feasible programs is maintained and undergoes an iterative code-level reproduction process. Those programs with better fitness are selected and kept until the end of evolution. However, unlike the heuristic search focused in prior works, BBO benchmark design presents unique challenges that require specialized solutions: 1) \textbf{Complex design objectives alignment}: We introduce a multi-objective program evolution paradigm that simultaneously optimizes the landscape similarity and algorithm distinguishing capability of benchmarks. 
2) \textbf{Program-landscape co-evolution}: We propose a novel reflection-based evolution strategy that leverages LLM's deep understanding of mathematical structures to achieve program-landscape co-evolution. By analyzing the causal relationship between program patterns and performance, this strategy enables the automatic synthesis of complex benchmark functions without explicit landscape editing. 

We summarize the contributions of this paper in threefold:
\begin{itemize}
    \item EoB represents the first paradigm shift from human-crafted to LLM-automated BBO benchmark design, delivering orders of magnitude efficiency improvement while maintaining benchmarking effectiveness. 
    \item Our key innovations include (i) the bi-objective program search modeling, (ii) the tailored LLM-based program evolution paradigm, and (iii) the reflective landscape editing scheme, which ensure the automation and performance of the search process. 
    \item EoB undergoes rigorous empirical validation across different perspectives: classic BBO algorithm benchmarking, modern learning-assisted optimizer training, and proxy modeling for expensive real-world problems, which underscores its universal applicability and practical value for the BBO community. 
\end{itemize}

\section{Related Works}\label{sec:related}
\subsection{BBO Benchmarks}
Compared to the exponential development scale of BBO algorithms, exploration and in-depth analysis of BBO benchmarks are still far from sufficient. Initial efforts focus on synthetic functions with closed forms, such as CoCo-BBOB~\cite{bbob2010}, IEEE CEC Benchmark suites~\cite{cec2021} and IOHProfiler~\cite{iohprofiler}. They are simple and easy to analyze, hence being commonly used in the community. Further efforts include category extensions~\cite{platemo,cec2017-constraint,mto-benchmark,dynamic-benchmark}, reproducible baselines~\cite{pypop7}, and efficiency optimization~\cite{evox}. However, recent advances in learning-assisted BBO algorithms~\cite{metabox,metabox-v2} reveal these synthetic benchmarks show limited performance evaluation coherence due to the gap between realistic tasks and these synthetic toy problems. Prior attempts to address this include hand-crafted extensions~\cite{proteindocking,hpob,uav,chip} or genetic programming-based search~\cite{GP-BBOB-1,GP-BBOB-2,GP-BBOB-3,lsre}. However, the former is both time- and resource-consuming, while the latter depends on deep expertise in domain-specific language. In contrast, Our EoB uses LLM-based program evolution to design benchmarks automatically and efficiently with minimal need for deep expertise.

\subsection{Program Evolution by LLMs}
We briefly review representative LLM-based program evolution methods. The core idea is that: 
by integrating LLMs with an evaluator to provide proper performance feedback, they might gain strength in open discovery through iterative in-context learning~\cite{icl}. The early efforts in the field includes FunSearch~\cite{funsearch} for heuristic discovery, EvoPrompting~\cite{evoprompting} for neural architecture search and EvoPromp~\cite{evoprompt} for automatic prompt optimization. They share an evolutionary paradigm where a population of feasible programs are maintained and go through tailored reproduce-then-select loop until an optimal program is evolved. Following these initial examples, a wide array of researches are emerged~\cite{eoh,reevo,llamea,eoh-moo,llm-moo,alphaevolve,eureka}. Given its prosperity, this paper presents the first exploration on using LLM-based program evolution for BBO benchmark design.

\section{Methodology}\label{sec:method}
\subsection{Overview}
Given $M$ design requirements~(objectives) $\mathcal{O}_1,...,\mathcal{O}_M$, such as the landscape diversity~\cite{ela-diversity-analysis} and algorithm distinguishing capability~\cite{algorithm-differential}, we formulate the benchmark design problem as a multi-objective optimization problem:
\begin{equation}\label{eq:probem-definition}
    \min_{f \in \mathbb{F}}(\mathcal{O}_1(f),...,\mathcal{O}_M(f)),
\end{equation}
where $\mathbb{F}$ spans a comprehensive function space, which in this paper are all feasible programs that could be used as objective functions for optimization problems. The $M$ design objectives are probably conflict with each other, hence we obtain a Pareto front set $\{f^*_1, ...,f^*_K\}$ with $K$ non-dominated program instances, which can be regarded as a benchmark suite with multiple diverse testing instances. For the basics of Pareto and domination, refer to~\cite{moo-paper-2}. 


\subsection{EoB}
EoB treats each benchmark function as an evolving program, searching through the vast space of possible mathematical formulations using a multi-objective evolutionary strategy in MOEA/D~\cite{moea/d} manner. It operates through five interconnected stages: initialization creates a diverse population of candidate functions; evaluation measures each function's landscape characteristics and algorithm discrimination power; evolution employs LLM-driven reflection to generate improved variants; selection maintains evolutionary pressure through Pareto-based comparison; and archive management preserves the best discoveries.

\subsubsection{Initialization}\label{sec:3.2.1}
The initialization in EoB resembles that of MOEA/D. First, we generate a group of reference vectors uniformly spread in the bi-objective space. The reference vectors are $N$ 2-dimensional ones:
\begin{equation}
    \vec{\lambda_i} = (\frac{i}{N-1},\frac{N-1-i}{N-1}), \quad i = 0,1,2,...,N-1.
\end{equation}
We then randomly sample a function program $f_i^{t=0}$ from LLM for each $\vec{\lambda_i}$, resulting in an initial solution population of size $N$, where $t$ denotes the evolutionary generation. To sample the program solutions from LLM, we repeatedly call an existing LLM using a carefully constructed \emph{Init\_Prompt}~(see Appendix~\ref{appx:prompts}). The aim of this prompt is, on the one hand, to help LLM grasp the basic context information, such as its role~(an optimization expert and benchmark designer) and its task~(to generate a program of an objective function). On the other hand, it should steer the LLM with sufficient expert-level specifications, which include two major parts.

\textbf{Generation Preference.} Given the bi-objective setting in EoB, we add an additional paragraph in \emph{Init\_Prompt} to inform the LLM which reference vector $\vec{\lambda_i}$ it generates a function program for. For example, when $\vec{\lambda_i} = (1,0)$, the LLM is instructed to generate a function program that considers maximizing its similarity with the target problem rather than its algorithm distinguishing capability.

\textbf{Population Diversity.} 
If we do not explicitly ask the LLM to generate function programs with different landscape characteristics, it tends to generate highly similar instances, which greatly harms the diversity in the initial population. To this end, we have summarized seven kinds of landscape construction knowledge from existing literature~\cite{landscape-1, landscape-2, deep-ela, ela}, and added them into our prompt to steer the initialization. We list three examples here and provide details of all seven types of knowledge in Appendix~\ref{appx:init_prompt}: 1) \emph{Asymmetric masking:} LLMs can use masking functions~(e.g., $\text{np.where}(x\leq b,\cdot,\cdot)$) to construct asymmetric objective landscapes; 2) \emph{Nonlinearity composition:} LLMs can use nested nonlinear operations~(e.g., $\frac{1}{\text{abs}(\text{tanh}(x)) + 1}$) to create highly twisted landscapes; 3) \emph{Polynomial periodicity:} LLMs can insert polynomial terms into periodic functions~(e.g., $\text{sin}(2\pi x^3)$) to create non-stationary periodic oscillations. During the initialization phase, we randomly select one of the seven types of landscape construction knowledge and add the selected one into the \emph{Init\_Prompt}, ensuring the initialized population comprises programs with diverse landscape characteristics.


\subsubsection{Evaluation}\label{sec:3.2.2}
While our EoB supports flexible design requirements (as shown in Eq.~\ref{eq:probem-definition}), we address two major objectives that are essential for BBO benchmark design in this paper: $\mathcal{O}_1$, Landscape Similarity Indicator (LSI), which indicates whether a searched program shows landscape similarity with the target hard-to-model BBO tasks; and $\mathcal{O}_2$, Algorithm Distinguishing Capability (ADC), which measures how effectively the searched program differentiates various BBO algorithms. Both objectives should be maximized.

\textbf{LSI.} Given a group of $H$ optimization tasks of interest $F_{target}:\{p_{1}, ..., p_{H}\}$, a primary objective is to design a group of $K$ synthetic testing instances $F:\{f_{1}, ..., f_{K}\}$ that exhibit similar landscape features to problems in $F_{target}$. Thus, we derive a similarity checking metric, LSI, to measure how similar a testing program $f\in\mathbb{F}$ designed by EoB is to $F_{target}$. Specifically, we first compute the landscape features of either problems in $F_{target}$ or the testing program $f$ through exploratory landscape analysis~(ELA)~\cite{ela}, which samples a set of solutions and corresponding objective values to compute diverse landscape features, such as convexity, for an optimization problem. Instead of directly using the computationally inefficient ELA features, we adopt NeurELA~\cite{neurela}, a recently proposed neural network-based landscape profiling system. For each $p \in F_{target}$, we use NeurELA to obtain their 16-dimensional feature vector $z_p$. For $f$, we obtain $z_f$. Then, the LSI of $f$ under $F_{target}$ is formulated as:
\begin{equation}\label{eq:LSI}
    \mathcal{O}_1(f|F_{target}) = \frac{1}{1 + \frac{1}{\sigma^2} \min_{p\in F_{target}}||z_f-z_p||_2}
\end{equation}
where $\sigma$ is a normalization factor to control the sensitivity of landscape feature similarity. A larger $\mathcal{O}_1$ indicates that the program $f$ designed by LLM is at least similar to a problem instance $p \in F_{target}$. This metric, by encouraging generated functions to align with various established landscape features, partly contributes to the overall diversity of the evolved benchmark suite. Further, by additionally considering the following ADC metric, EoB explores a set of solutions (benchmark programs) along the Pareto front, which exhibit good diversity.

\textbf{ADC.} The second benchmark design objective we consider in EoB is its algorithm distinguishing capability. Suppose we have a testing program $f$ designed by EoB and a BBO algorithm pool $\Lambda:\{A_1,...,A_L\}$ with $L$ BBO algorithms. ADC measures performance variation across these BBO algorithms on $f$ by first running the $L$ algorithms on $f$ for $B$ random seed-controlled independent runs. We record a $B\times L$-dimensional objective matrix $Obj$ where each entry $Obj_{i,j}$ records the best-so-far objective value achieved by $A_j$ while optimizing $f$ at the $i$-th independent run. Then the ADC of $f$ over the algorithm pool $\Lambda$ is formulated as:
\begin{equation}\label{eq:ADC}
    \mathcal{O}_2(f|\Lambda) = \text{std}(\frac{1}{B}\sum_{i=1}^{B}\vec{Obj}_{i,:}),
\end{equation}
where $\vec{Obj}_{i,:}$ denotes the best-so-far objective values achieved by all BBO algorithms in $\Lambda$ at the $i$-th run. We also note that to address the variation in objective values across different $f$ designed by EoB, we apply min-max normalization on the $Obj$ matrix. A larger ADC indicates that $f$ is sufficiently good and challenging for distinguishing diverse BBO algorithms in terms of their performances.

\textbf{PBI.} For each individual program $f_i$ in the initialized population and subsequent evolving populations, we first compute its LSI $\mathcal{O}_1(f)$ and ADC $\mathcal{O}_2(f)$. After objective computation, we further compute Penalty-based Boundary Intersection~(PBI), a scalarization trick to turn multi-objective space into measurable scalar performance. PBI receives three inputs: $\mathcal{O}_1(f_i)$, $\mathcal{O}_2(f_i)$, and the reference vector $\vec{\lambda}_i$. It then outputs a scalar score that reflects the aggregated performance of $f_i$ between the tradeoff of LSI~($\mathcal{O}_1$) and ADC~($\mathcal{O}_2$). The PBI score is used as reference information for subsequent LLM-based program evolution, including both the reproduction~(Sec.~\ref{sec:3.2.3}) and selection~(Sec.~\ref{sec:3.2.4}) phases. For details of PBI's computation, refer to the original MOEA/D paper~\cite{moea/d}.

\subsubsection{Evolution}\label{sec:3.2.3}
Given a population of candidate programs $\{f_i^{t}\}_{i=1}^{N}$ at the $t$-th generation (~$t=0$ is the initialized population), we propose a two-stage reproduction scheme to fully leverage the LLM's outstanding reasoning capability for code-level understanding and program search. 

\textbf{Reflection stage.} At the first stage, we iteratively prompt the LLM with pairs of candidate programs and instruct it to compare the objective scores of these program pairs. The LLM then engages in comparison and performs reflective thinking, resulting in valuable insights on how to improve the candidate programs. We provide a detailed prompt for this stage in Appendix~\ref{appx:reflection}. Specifically, similar to MOEA/D, we first construct the neighborhood set for each $f_i^{t}$, which comprises 5 candidate programs with the closest Euclidean distance to $f_i^{t}$ in terms of the corresponding reference vectors. Then, we randomly select two candidates $f^{\prime}$ and $f^{\prime\prime}$ from the neighborhood set. 
For each candidate program, the LLM is provided with three inputs: (i) Source code, (ii) scores of LSI, ADC, and PBI, and (iii) reference vectors. Considering that the candidate program is generated by the LLM, it may encounter syntax errors and other invalid issues; therefore, EoB steers the LLM to output comprehensive analyzes on a case-by-case basis: 
\begin{itemize}
    \item \textbf{Aggressive Mutation:} If $f^{\prime}$ and $f^{\prime\prime}$ are both invalid, the LLM is directed to analyze the reasons behind them and output corresponding suggestions to rectify the invalid issues or facilitate a complete regeneration.
    \item \textbf{Conservative Mutation:} If one of them is invalid, the LLM is directed to ignore the invalid one and focus on the valid one to analyze why it works in LSI/ADC/PBI aspects and subsequently propose the corresponding improvement suggestions. 
    \item \textbf{Reflection-Based Crossover:} If both of them are valid, the LLM is directed to conduct a three-fold in-depth reflection when comparing $f^{\prime}$ with $f^{\prime\prime}$~(where the candidate with the higher PBI is designated as the \emph{winner} and the other as the \emph{loser}): (i) Successful Pattern Identification: recognize the major code-level differences between \emph{ winner} and \emph{loser}, summarize knowledge and its relationship with landscape characteristics. (ii) Beneficial Gene Preservation: if the \emph{loser} shows a higher score in either LSI or ADC, the LLM is asked to analyze the reason behind it and preserve such valuable knowledge. (iii) Bi-objective Tradeoff Analysis: recognize and analyze the positive/negative impacts of certain parts of the programs' codes on LSI and ADC, with a special focus on the conflict case such as adding a mathematical operation increase LSI but harm ADC. 
\end{itemize}
In the end of this stage, all intermediate analyzes and thoughts are summarized by the LLM and organized in a JSON-like response, including a summary of the overall reasoning process, condensed analysis results for the aforementioned case-by-case reflections, and suggestions/instructions for generating the next generation offspring program.

\textbf{Reproduction stage.} Given the JSON-like reflection obtained in the Reflection stage, the LLM is directed to refer to and follow these useful analysis or improvement suggestions to produce a single new program candidate $f_i^{t+1}$. By repeatedly calling this two-stage evolution scheme $N$ times, we could obtain an updated candidate population. We provide detailed prompts for this
stage at Appendix~\ref{appx:reproduction} Such comparison-based reflection and construction could facilitate fine-grained and effective benchmark function discovery through the co-evolution of both code-level programs and corresponding landscape-level characteristics.

\subsubsection{Selection}\label{sec:3.2.4}
Given the parent population $\{f_i^{t}\}_{i=1}^{N}$ and the newly reproduced offspring population $\{f_i^{t+1}\}_{i=1}^{N}$, we first compare the PBI scores between $f_i^{t}$ and $f_i^{t+1}$; the one with the higher score survives. To further enhance exploration and exploitation in the evolutionary process, we additionally allow comparison between $f_i^{t+1}$ and the neighbors of $f_i^{t}$. A neighboring parent program will be replaced by $f_i^{t+1}$ if its PBI is lower. We restrict such neighboring selection operations to 2 times per individual to avoid premature convergence. 

\subsubsection{Pareto Front Maintenance}\label{sec:3.2.5}
EoB maintains the up-to-date Pareto front set along the iterative program evolution. In this work, we define Pareto dominance between two programs $f_i$ and $f_j$ as: $f_i$ dominates $f_j$ if and only if $\mathcal{O}_1(f_i) \geq \mathcal{O}_1(f_j)$ and $\mathcal{O}_2(f_i) \geq \mathcal{O}_2(f_j)$, and vise versa. A program is regarded as a Pareto front if all other programs in the population cannot dominate it. Initially, an empty Pareto front set $PF$ is established. After initialization, $PF$ is updated by the non-dominated programs in the population. At subsequent generations, $PF$ is updated by first being concatenated with the selected non-dominated elite programs (from the union of parents and offspring), and then retaining those non-dominated ones.

\begin{algorithm}[t]
\caption{Evolution of Benchmark~(EoB)}
\label{alg:algorithm}
\textbf{Input}: Large Language Model $\mathcal{LLM}$, Target BBO tasks $F_{target}$, BBO algorithm pool $\Lambda$, \#Reference vectors $N$, \#Evolution generations $T$, Pareto set $PF=\emptyset$.\\
\textbf{Output}: Finally obtained benchmark 
\begin{algorithmic}[1] 
\STATE Let $t=0$, initialize population $\{f_i^{t}\}_{i=1}^{N}$~(Sec.~\ref{sec:3.2.1});
\STATE Evaluate each $f_i^t$~(Sec.~\ref{sec:3.2.2}), update $PF$~(Sec.~\ref{sec:3.2.5});
\WHILE{$t < T$}
\STATE Reproduce offspring $\{f_i^{t+1}\}_{i=1}^{N}$~(Sec.~\ref{sec:3.2.3});
\STATE Evaluate each $f_i^{t+1}$~(Sec.~\ref{sec:3.2.2});
\STATE Select elites from $\{f_i^{t}\}_{i=1}^{N} \cup \{f_i^{t+1}\}_{i=1}^{N}$~(Sec.~\ref{sec:3.2.4});
\STATE Update $PF$~(Sec.~\ref{sec:3.2.5}) with elites, $t \leftarrow t+1$;
\ENDWHILE
\STATE \textbf{return} $PF$
\end{algorithmic}
\end{algorithm}

After all designs have been introduced above, we present the overall workflow of EoB in Alg.~\ref{alg:algorithm}, which shows a simple pipeline. After the evolution, each program in the Pareto front set can be regarded as a good benchmark function. The whole set can hence be regarded as a complete benchmark, robustly providing both desired problem landscape diversity and effective algorithm distinguishing capabilities..

\begin{figure*}[t]
\centering
\includegraphics[width=0.95\linewidth]{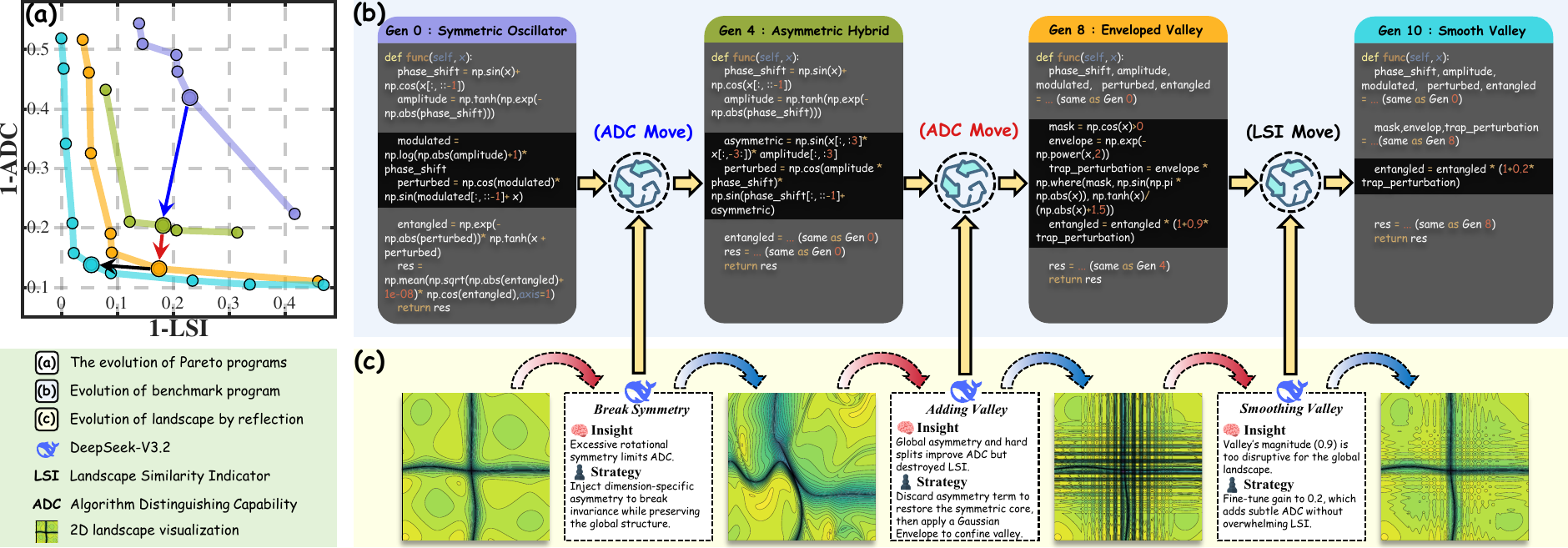}
\caption{Visualization of EoB on the co-evolution of the benchmark program and corresponding landscape characteristics.}
\label{fig:4.2.1}
\end{figure*}

\section{Experimental Results}
In this section, we provide empirical validation to demonstrate 1) EoB provides an automated way to address  multi-dimensional benchmarking needs in BBO community; 2) Each design proposed for EoB is effective. In the rest of this section, we first summarize basic experimental settings in Sec.~\ref{sec:4.1} and then dive into the above two aspects in Sec.~\ref{sec:4.2}, Sec.~\ref{sec:4.3} and Sec.~\ref{sec:4.4}.

\subsection{Basic Setup}\label{sec:4.1}
\textbf{EoB.} The base LLM used in EoB is Deepseek-v3.2~\cite{deepseek-v3.2} with its default setting. For all of the following experiments, we set the number of reference vectors $N=32$~(which is also the population size), the number of evolution generations $T=10$. As for the BBO algorithm pool $\Lambda$, we select from related literature 10 representative BBO algorithms: Vanilla DE~\cite{de}, JDE21~\cite{jDE21}, MadDE~\cite{MadDE}, NL-SHADE-LBC~\cite{nlshadelbc}, PSO~\cite{pso}, GLPSO~\cite{glpso}, sDMS-PSO~\cite{sdms-pso}, SAHLPSO~\cite{sahl-pso}, CMA-ES~\cite{cmaes}, Sep-CMA-ES~\cite{sepcmaes}. They cover diverse categories and show diverse optimization capabilities in our preliminary study, hence beneficial for measuring ADC score. We use these $L=10$ algorithms and $B=10$ independent runs to compute ADC scores. The weighted sum strength factor $\theta$ in PBI score computation is set to 5. At last, we note that the choice of target BBO tasks $F_{target}$ varies in different experiments to demonstrate multi-dimensional usages of EoB, which we will detail in each experiment section.

\textbf{Involved Benchmarks.} Several benchmarks are involved to serve as both the compared baselines and the target BBO tasks $F_{target}$: 1) CoCo-BBOB~\cite{bbob2010,hansen2021coco}: which includes 24 hand-crafted synthetic functions with diverse landscape characteristics. It is the most representative BBO benchmark used by optimization researchers to evaluate and compare their BBO algorithms. 
2) MetaBox~\cite{metabox,metabox-v2}, a benchmark platform proposed recently for learning-based BBO algorithms~\cite{ma2025toward}. It contains a wide array of realistic benchmarks such as UAV path planning~\cite{uav}~(56 testing cases), hyper-parameter optimization~\cite{hpob}~(935 testing cases). We use UAV, HPO to denote them. All experiments are conducted on a platform equipped with an RTX 2080Ti 11GB GPU, an Intel Xeon E5-2680 v4 @ 56x 3.3GHz CPU, and 128GB RAM. We provide our project at \url{https://anonymous.4open.science/r/EoB}.

\subsection{Benchmarking BBO algorithms}\label{sec:4.2}
One of the primary usages of EoB is to extend~(construct) from representative benchmarks to enhance their benchmarking capability, such as the algorithm distinguishing capability. To this end, we set the target BBO tasks $F_{target}$ as CoCo-BBOB benchmark, and run EoB~(Alg.~\ref{alg:algorithm}) multiple times to obtain 24 10-dimensional function programs\footnote{One EoB run can not ensure the finally obtained $PF$ contains 24 non-dominated programs, multiple runs~(2-3) are needed.}~(the same size as CoCo-BBOB), which we name the searched benchmark as EoB-BBOB. We leave the detailed function programs in EoB-BBOB at our \href{https://anonymous.4open.science/r/EoB/result/EoB-BBOB.py}{project}. We provide following interesting results to present how well EoB performs.

\subsubsection{Evolution Effectiveness}
We first illustrate the evolution path of EoB's program evolution process in Fig.~\ref{fig:4.2.1}. In the left of this figure, we plot the pareto front improvement during the evolution, which shows that the programs evolved by EoB continuously improve both the LSI and ADC scores along the program evolution, clearly demonstrating the effectiveness of EoB. We further delve into the details of such effective evolution in the right of this figure, where we showcase the concrete evolution path of a program individual~(from its initialization to the end of search). The results there validate a key design motivation we mentioned before: by using LLM-based program evolution, EoB is capable of evolving the landscape of benchmark functions and its code-level implementation simultaneously. Driven by our proposed LLM-based bi-objective evolution, the function program is evolved step by step with clear and interpretable design thoughts. This significantly mitigates the expert-level knowledge and labor input required to design a benchmark. 

\subsubsection{Benchmark Diversity Comparison}
\begin{figure}[t]
\centering
\includegraphics[width=0.9\linewidth]{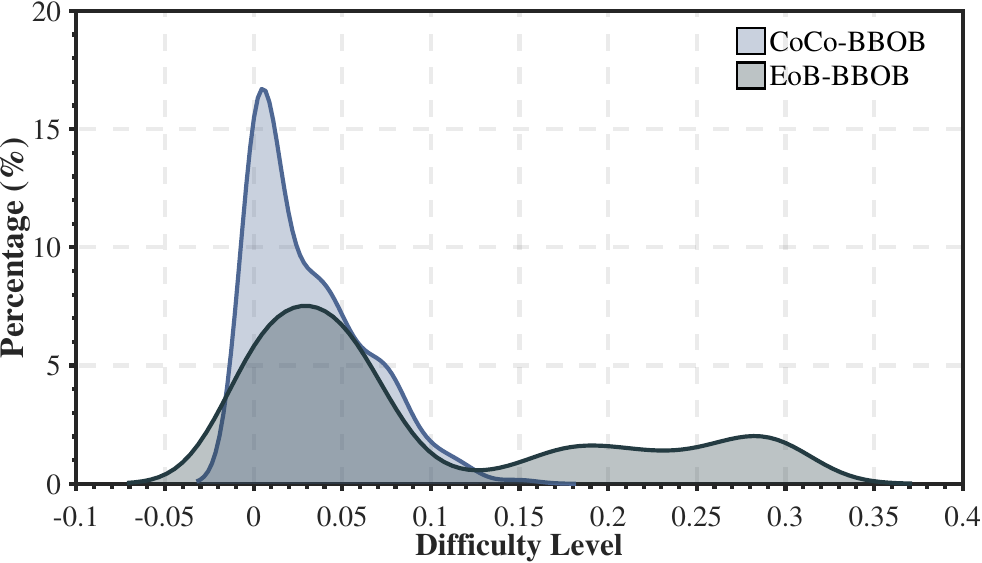}
\caption{Benchmark diversity of CoCo-BBOB and EoB-BBOB.}
\label{fig:4.2.2}
\end{figure}

We further illustrate the instance-level diversity~(more uniform is better) of CoCo-BBOB and the searched EoB-BBOB in Fig.~\ref{fig:4.2.2}. The results there are obtained by first testing the 10 algorithms in $\Lambda$ on each of the 24 testing instances in a benchmark for 10 independent runs. For $i$-th testing instance and $j$-th run, we record the standard deviation of best-so-far objective values~(min-max normalized) found by the 10 algorithms. A larger standard deviation indicates the testing instance is more challenging and hence  easier to distinguish different BBO algorithms. Then, we obtain 240 standard deviation values which we use to plot the distribution density in Fig.~\ref{fig:4.2.2}. The results may indicate that: 1) Almost all hand-crafted synthetic functions in CoCo-BBOB show low algorithm distinguishing capability and difficulty diversity; 2) EoB could automatically design a more uniform and diversified benchmark by LLM-based program evolution, which is promising.

\subsection{Training Learnable Optimizer}\label{sec:4.3}
In this section we explore whether EoB could also address the new benchmarking needs in our community. More recently, learning-based BBO algorithms~\cite{rl-bbo-survey,metabbo-survey-yx,ma2025toward,ml-bbo-survey} emerge as an active and promising research avenue. They address the potential limitations in human-crafted BBO algorithms such as the adaptability issue and limited generalization by meta-learning~\cite{meta-learning,meta-learning-1} effective algorithm control policy that boosts classical BBO algorithms across a problem distribution. The finally learned BBO algorithms hence show preferred generalization towards unseen BBO tasks. Such paradigm gives birth to a large array of learnable optimizers~\cite{RLDAS,ALDes,metaes,wu2025learning,q-mamba,chengran-metabbo,du-metabbo,tianye-metabbo,xueke-metabbo,wukai-metabbo,pei2025libog}. In this experiment, to ensure fast implementation and rigorous experimental protocol, we use four MetaBBO baselines well-maintained in MetaBox~\cite{metabox,metabox-v2}: LDE~\cite{LDE}, GLEET~\cite{GLEET}, SYMBOL~\cite{Symbol} and RLDEAFL~\cite{RLDE-AFL}.  

\begin{figure}[t]
\centering
\includegraphics[width=0.98\linewidth]{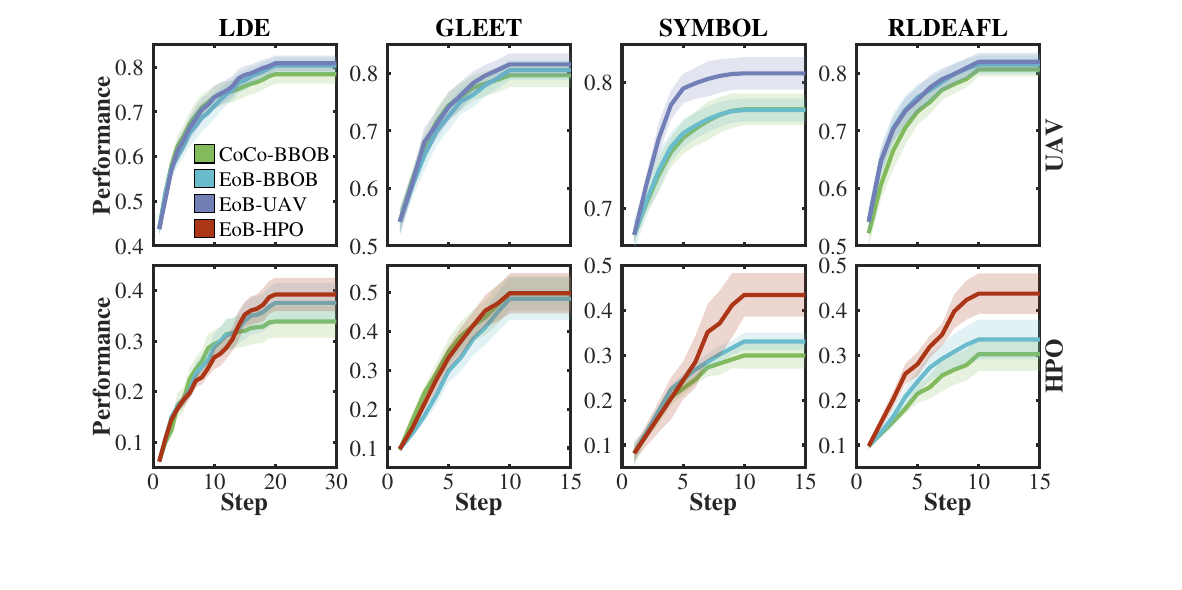}
\caption{Zero-shot generalization performances of four learnable optimizers when trained by different benchmarks. Testing performance curves are presented.}
\label{fig:4.3.1}
\vspace{-3mm}
\end{figure}

\subsubsection{Generalization Advantage}
The most important aspect cared about by the users of learnable optimizers is its true generalization potential. To validate EoB's advantage on this aspect, we establish two validation scenarios: 1) A user trains learnable optimizer on a benchmark and then expects the trained model generalizes well on UAV tasks; 2) The same scenario yet the target tasks are HPO tasks. The benchmark baselines used for training in 1) are CoCo-BBOB, EoB-BBOB we searched in Sec~\ref{sec:4.2} and EoB-UAV, which is searched by EoB using UAV-train-set\footnote{Benchmarks in MetaBox are split into a train/test set to support learnable optimizers.} as the $F_{target}$ in Alg.~\ref{alg:algorithm}. And the baselines for 2) are CoCo-BBOB, EoB-BBOB and EoB-HPO~(HPO-train-set as $F_{target}$). For the four learnable optimizer baselines LDE, GLEET, SYMBOL and RLDEAFL, we train them on the two scenarios by their default settings and the three benchmark baselines in the two scenarios respectively, and test the trained models on UAV-train-set and HPO-train-set for 10 independent optimization episodes. We plot the optimization curves in terms of generalization performances in Fig.~\ref{fig:4.3.1}, where the results clearly underscore that: 1) Through the bi-objective program evolution, the benchmark optimized by EoB could provide high-quality training source for learnable optimizers; 2) CoCo-BBOB, while regarded as a golden standard for evaluating classic BBO algorithms, may not be a proper choice for learning-assisted methods due to its biased human design. However, up to now, it is still the first choice in many learnable optimizers. The searched programs in EoB-UAV/HPO are at our \href{https://anonymous.4open.science/r/EoB/result/EoB-UAV.py}{project}.

We also want to emphasize the advantage of EoB in the efficiency side. Realistic BBO task such as UAV and HPO we discussed above are often quite expensive~(either time-consuming or resource-consuming). However, training an advanced learnable optimizer require repeated interplay with the target BBO tasks. If the target BBO tasks are expensive, training on them is impractical. To demonstrate this aspect, we present at Table~\ref{tab:time_comparison} the average time consumed by the four learnable optimizer baselines per learning step when they use UAV/HPO as training set directly or use EoB-UAV/EoB-HPO instead. A significant efficiency leap could be observed from the results, which highlights an additional usage of EoB: searching for proxy benchmark for those expensive and hard-to-model BBO tasks.


\begin{table}[t]
\centering
\caption{Evaluation efficiency comparison.}
\label{tab:time_comparison}
\footnotesize
\resizebox{0.8\linewidth}{!}{ 
\begin{tabular}{l|c|cccc} 
\toprule
\multicolumn{2}{c|}{\multirow{2}{*}{\diagbox[width=10em]{\textbf{Algorithm}}{\textbf{Scenario}}}} & 
\multicolumn{2}{c}{\textbf{UAV}} & 
\multicolumn{2}{c}{\textbf{HPO}} \\
\cmidrule(lr){3-4} \cmidrule(lr){5-6}

\multicolumn{2}{c|}{} & Real & EoB & Real & EoB \\ 
\midrule

\multirow{2}{*}{LDE} 
 & Normal & 30.38 min & \textbf{3.124 s} & 23.72 min & \textbf{2.923 s}\\
 & Large   & $\approx$ 7 h & \textbf{28.32 s} & $\approx$ 4 h & \textbf{27.43 s} \\
\midrule

\multirow{2}{*}{GLEET} 
 & Normal & 26.23 s & \textbf{0.256 s}& 14.68 s & \textbf{0.231 s}\\
 & Large  & 284.2 s & \textbf{0.287 s} & 146.6 s & \textbf{0.274 s} \\
\midrule

\multirow{2}{*}{SYMBOL} 
 & Normal & 43.68 s & \textbf{0.423 s}& 190.8 s & \textbf{0.412 s} \\
 & Large  & 78.95 s & \textbf{0.528 s} & 312.1 s & \textbf{0.547 s} \\
\midrule

\multirow{2}{*}{RLDEAFL}
 & Normal & 98.24 s & \textbf{0.921 s} & 15.98 s & \textbf{0.852 s} \\
 & Large  & 22.91 min & \textbf{1.442 s} & 205.7 s & \textbf{1.352 s} \\

\bottomrule
\end{tabular}} 
\end{table}

\vspace{-3mm}

\begin{table}[t]
\centering
\caption{GLHF Performance}
\label{tab:glhf_perf}
\footnotesize
\resizebox{0.7\linewidth}{!}{
\begin{tabular}{l|ccc}
\toprule
\rowcolor[HTML]{E7E7E7}
\textbf{Scenario} & \textbf{CoCo-BBOB} & \textbf{EoB-BBOB} & \textbf{EoB-Scenario} \\
\midrule
\multirow{2}{*}{UAV}
 & 8.380E-01 & 7.508E-01 & \textbf{8.484E-01} $\uparrow$\\
 & $\pm$5.193E-02 & $\pm$1.149E-01 & $\pm$\textbf{3.854E-02} \\
\midrule
\multirow{2}{*}{HPO}
 & 5.144E-01 & 5.214E-01 & \textbf{6.580E-01} $\uparrow$\\
 & $\pm$2.113E-01 & $\pm$1.769E-01 & $\pm$\textbf{1.933E-01} \\
\bottomrule
\end{tabular}
}
\end{table}


\subsubsection{Gradient-dependent Case}
Another interesting usage of EoB is when we consider a specific kind of learnable optimizer that requires the gradient information of the training problems to supervise its neural network model. An example is GLHF~\cite{GLHF}. This specific feature hinders such learnable optimizers from direct training on realistic BBO tasks where gradient information is not accessible, e.g., UAV/HPO. Fortunately, EoB could address this issue by searching executable function programs that show both  high similarity~(LSI) with the target tasks and substantial difficulty diversity~(ADC), e.g., EoB-UAV/HPO. More importantly, they can be easily transformed into any autograd framework such as \emph{torch} and \emph{jax}, by LLM. The results in Table~\ref{tab:glhf_perf} validate EoB's advantage for GLHF-like learnable optimizers.  


\begin{table}[t]
\centering
\caption{Performance Consistency}
\label{tab:consistency}
\footnotesize
\resizebox{0.75\linewidth}{!}{
\begin{tabular}{l|ccc}
\toprule
\rowcolor[HTML]{E7E7E7}
\textbf{Scenario} & \textbf{CoCo-BBOB} & \textbf{EoB-BBOB} & \textbf{EoB-Scenario} \\
\midrule
UAV & 0.003 & 0.396 & \textbf{0.453} $\uparrow$ \\
HPO & 0.694 & 0.702 & \textbf{0.761} $\uparrow$ \\
\bottomrule
\end{tabular}
}
\end{table}

\subsubsection{Key Reason Behind}
At last, we give a simple yet intuitive interpretation on the reason behind EoB's strength. Specifically, we first define a simple metric $\mathbb{C}(B_{source},B_{target})$ that measures the performance consistency between a source benchmark $B_{source}$ and a target benchmark $B_{target}$. Specifically, we first test the 10 BBO algorithms in our proposed algorithm pool $\Lambda$ on $B_{source}$ to obtain a 10-dimensional performance vector $\vec{S}_{source}:\{\frac{1}{rank_i}\}_{i=1}^{10}$, where $rank_i$ is the average rank of $i$-th algorithm in $\Lambda$. We use the same logic to obtain $\vec{S}_{target}$. Then the performance  consistency is computed as:
\begin{equation}
    \mathbb{C}(B_{source},B_{target}) = 1 - \mathcal{W}(\vec{S}_{source}, \vec{S}_{target}), 
\end{equation}
where $\mathcal{W}(\cdot,\cdot)$ denotes the Wasserstein distance. A larger consistency score indicates that the performance of BBO algorithms on the source benchmark is more consistent with the performance on the target task. Such performance predictivity may support the generalization of learnable optimizers. To validate this, we present in Table~\ref{tab:consistency} the performance consistency between \{CoCo-BBOB, EoB-BBOB, EoB-UAV, EoB-HPO\} and \{UAV, HPO\}. The results there, combined with the results in Fig.~\ref{fig:4.3.1} and Table~\ref{tab:glhf_perf} cross-validate what we have assumed.


\begin{figure}[t]
\centering
\includegraphics[width=\linewidth]{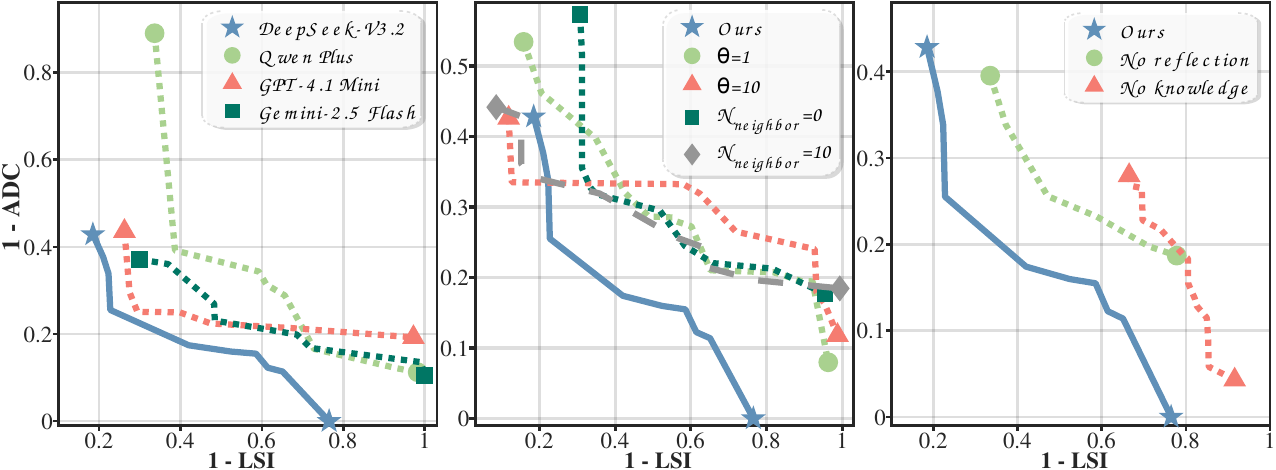}
\caption{The ablation results shown as the pareto front set $PF$ found by different baselines.}
\label{fig:4.4.1}
\end{figure}
\vspace{-3mm}

\subsection{Ablation Studies}\label{sec:4.4}
To validate the contribution of each component within EoB, we conducted three ablation studies: 1) LLM Backbone Capability, assessing the impact of the underlying model on evolutionary efficiency; 2) Hyperparameter Sensitivity, analyzing the effects of neighborhood size ($N_{neighbor}$) and PBI penalty ($\theta$) on multi-objective optimization; and 3) Key Component Analysis, removing the Reflection Stage and Landscape Initialization Knowledge. Using the UAV-train set as the target task, the resulting Pareto fronts (Fig.~\ref{fig:4.4.1}) reveal that: 1) Deepseek-v3.2 achieves the most balanced performance among the backbones. 2) Diversity-oriented settings ($\theta=10, N_{neighbor}=10$) enhance LSI but compromise the objective trade-off, whereas exploitation-heavy settings ($\theta=1, N_{neighbor}=1$) suffer from low efficiency due to restricted information exchange. 3) The absence of Landscape Knowledge causes LSI stagnation, while removing the Reflection Stage degrades overall efficiency by severing the feedback loop of comparative insights. 

\section{Discussion}
\textbf{Takeaways.} 1) If you are seeking a more challenging synthetic benchmark to test your BBO algorithms, feel free to use our EoB-BBOB; 2) If you are developing a learning-assisted BBO algorithm to solve some kind of expensive realistic tasks, run EoB to attain cheap synthetic benchmark to save your time and money; 3) You may consider using EoB to assist your analysis on some hard-to-model BBO tasks by extending their proxy.

\textbf{Future Directions.} We would like to discuss a few promising directions. The first is extending EoB towards diverse optimization categories, e.g., constrained optimization~\cite{cec2017-constraint}, multi-objective optimization~\cite{cec-moo} and dynamic optimization~\cite{dynamic-benchmark}, for which EoB provides a universal and automated pipeline. Another important direction is to make the benchmark design objectives more fine-grained. While two design objectives we considered in this paper more or less provide an effective way to ensure comprehensive benchmarking, additional design objectives such as computational complexity~\cite{evox} and surrogate difficulty~\cite{Surr-RLDE} could definitely help benchmarking in complex optimization tasks. At last, one may also consider using EoB's paradigm for research fields such as AutoML~\cite{automal-bench}, AI4Science~\cite{agentsci} and Benchmark of Benchmark~\cite{benchofbench}, however, more efforts and certain expertise are still needed for adaption. 

\section{Conclusion}
In this work, we address the bottleneck in benchmark design for the BBO field by replacing the laborious hand-crafted design process with an automatic paradigm. To this end, we propose EoB, the first systematic framework to facilitate the evolution of high-quality benchmarks. EoB operates under a bi-objective LLM-based program search, aiming to discover evaluation programs~(BBO instances) that simultaneously exhibit problem landscape diversity and algorithm distinguishing capabilities. EoB is fully automatic through iterative multi-turn LLM conversations, ensuring minimal human-level design bias. Beyond its automation, EoB's end-to-end paradigm supports diverse needs, including synthetic benchmark construction, training set design for learning-based BBO algorithms and proxy extension for hard-to-model BBO tasks. 
Utimately, this work not only introduces a powerful new tool for automatic benchmark generation, but it also highlights the indispensable role of thoughtfully designed evaluation landscapes in advancing BBO research.

\section*{Impact Statement}
We claim a significant impact of our work on both research development in the optimization community and, more importantly, its foundational impact on how researchers and engineers approach the evaluation, refinement, and redesign of optimization algorithms. By automating the benchmark design process, the substantial labor resources and specialized expertise costs could be significantly reduced. Such an automated pipeline also helps mitigate human-based bias to assure benchmarking objectivity. This paradigm shift will significantly transform the evaluation and refinement of optimization algorithms, enabling more rigorous and objective comparisons that accelerate advancements across BBO and its diverse applications

\bibliography{ref}

@article{bbo-survey,
  title={Efficient global optimization of expensive black-box functions},
  author={Jones, Donald R and Schonlau, Matthias and Welch, William J},
  journal={Journal of Global optimization},
  volume={13},
  number={4},
  pages={455--492},
  year={1998},
  publisher={Springer}
}

@book{engineer1,
  title={Genetic algorithms and engineering optimization},
  author={Gen, Mitsuo and Cheng, Runwei},
  year={1999},
  publisher={John Wiley \& Sons}
}

@article{scientific1,
  title={Black-box optimization for automated discovery},
  author={Terayama, Kei and Sumita, Masato and Tamura, Ryo and Tsuda, Koji},
  journal={Accounts of Chemical Research},
  volume={54},
  number={6},
  pages={1334--1346},
  year={2021},
  publisher={ACS Publications}
}

@article{es-llm,
  title={Evolutionary System 2 Reasoning: An Empirical Proof},
  author={Ma, Zeyuan and Huang, Wenqi and Song, Guo-Huan and Guo, Hongshu and Ma, Sijie and Cao, Zhiguang and Gong, Yue-Jiao},
  journal={arXiv preprint arXiv:2512.05760},
  year={2025}
}

@book{1992ga,
  title={Adaptation in natural and artificial systems: an introductory analysis with applications to biology, control, and artificial intelligence},
  author={Holland, John H},
  year={1992},
  publisher={MIT press}
}

@article{2002es,
  title={Evolution strategies--a comprehensive introduction},
  author={Beyer, Hans-Georg and Schwefel, Hans-Paul},
  journal={Natural Computing},
  year={2002},
}

@inproceedings{pso,
  title={Particle swarm optimization},
  author={Kennedy, James and Eberhart, Russell},
  booktitle={Proceedings of ICNN'95-International Conference on Neural Networks},
  year={1995}
}

@article{de,
  title={Differential evolution-a simple and efficient heuristic for global optimization over continuous spaces},
  author={Storn, Rainer and Price, Kenneth},
  journal={Journal of Global Optimization},
  year={1997},
}

@inproceedings{meta-learning,
  title={Model-agnostic meta-learning for fast adaptation of deep networks},
  author={Finn, Chelsea and Abbeel, Pieter and Levine, Sergey},
  booktitle={ICML},
  year={2017}
}

@incollection{meta-learning-1,
  title={Learning to learn: Introduction and overview},
  author={Thrun, Sebastian and Pratt, Lorien},
  booktitle={Learning to learn},
  year={1998},
}

@article{ma2025toward,
  title={Toward automated algorithm design: A survey and practical guide to meta-black-box-optimization},
  author={Ma, Zeyuan and Guo, Hongshu and Gong, Yue-Jiao and Zhang, Jun and Tan, Kay Chen},
  journal={IEEE TEVC},
  year={2025},
}

@inproceedings{metabox,
    author={Ma, Zeyuan and Guo, Hongshu and Chen, Jiacheng and Li, Zhenrui and Peng, Guojun and Gong, Yue-Jiao and Ma, Yining and Cao, Zhiguang},
    title={MetaBox: A Benchmark Platform for Meta-Black-Box Optimization with Reinforcement Learning},
    booktitle = {NeurIPS},
    year={2023},
}

@inproceedings{
neurela,
title={Neural Exploratory Landscape Analysis for Meta-Black-Box-Optimization},
author={Zeyuan Ma and Jiacheng Chen and Hongshu Guo and Yue-Jiao Gong},
booktitle={ICLR},
year={2025},
}

@article{hansen2021coco,
  title={COCO: A platform for comparing continuous optimizers in a black-box setting},
  author={Hansen, Nikolaus and Auger, Anne and Ros, Raymond and Mersmann, Olaf and Tu{\v{s}}ar, Tea and Brockhoff, Dimo},
  journal={Optimization Methods and Software},
  year={2021},
}

@phdthesis{bbob2010,
  title={Real-parameter black-box optimization benchmarking 2010: Experimental setup},
  author={Hansen, Nikolaus and Auger, Anne and Finck, Steffen and Ros, Raymond},
  year={2010},
  school={INRIA}
}

@inproceedings{iohprofiler,
  title={Benchmarking discrete optimization heuristics with IOHprofiler},
  author={Doerr, Carola and Ye, Furong and Horesh, Naama and Wang, Hao and Shir, Ofer M and B{\"a}ck, Thomas},
  booktitle={GECCO},
  year={2019}
}

@techreport{cec2021,
  title={Problem definitions and evaluation criteria for the {CEC} 2021 Special Session and Competition on Single Objective Bound Constrained Numerical Optimization},
  author={Ali Wagdy Mohamed and Anas A Hadi and Ali Khater Mohamed and Prachi Agrawal and Abhishek Kumar and P. N. Suganthan},
  year={2021},
}

@article{metabox-v2,
  title={MetaBox-v2: A Unified Benchmark Platform for Meta-Black-Box Optimization},
  author={Ma, Zeyuan and Gong, Yue-Jiao and Guo, Hongshu and Qiu, Wenjie and Ma, Sijie and Lian, Hongqiao and Zhan, Jiajun and Chen, Kaixu and Wang, Chen and Huang, Zhiyang and others},
  journal={arXiv preprint arXiv:2505.17745},
  year={2025}
}

@article{RLDAS,
  title={Deep Reinforcement Learning for Dynamic Algorithm Selection: A Proof-of-Principle Study on Differential Evolution},
  author={Guo, Hongshu and Ma, Yining and Ma, Zeyuan and Chen, Jiacheng and Zhang, Xinglin and Cao, Zhiguang and Zhang, Jun and Gong, Yue-Jiao},
  journal={TSMC},
  year={2024},
}

@inproceedings{jDE21,
  title={Self-adaptive differential evolution algorithm with population size reduction for single objective bound-constrained optimization: Algorithm j21},
  author={Brest, Janez and Mau{\v{c}}ec, Mirjam Sepesy and Bo{\v{s}}kovi{\'c}, Borko},
  booktitle={CEC},
  year={2021},
}

@inproceedings{MadDE,
  title={Improving differential evolution through Bayesian hyperparameter optimization},
  author={Biswas, Subhodip and Saha, Debanjan and De, Shuvodeep and Cobb, Adam D and Das, Swagatam and Jalaian, Brian A},
  booktitle={CEC},
  year={2021}
}

@article{LDE,
  title={Learning adaptive differential evolution algorithm from optimization experiences by policy gradient},
  author={Sun, Jianyong and Liu, Xin and B{\"a}ck, Thomas and Xu, Zongben},
  journal={IEEE TEVC},
  year={2021},
}

@inproceedings{GLEET,
  title={Auto-configuring Exploration-Exploitation Tradeoff in Evolutionary Computation via Deep Reinforcement Learning},
  author={Ma, Zeyuan and Chen, Jiacheng and Guo, Hongshu and Ma, Yining and Gong, Yue-Jiao},
  booktitle={GECCO},
  year={2024}
}

@inproceedings{Symbol,
  title={Symbol: Generating Flexible Black-Box Optimizers through Symbolic Equation Learning},
  author={Chen, Jiacheng and Ma, Zeyuan and Guo, Hongshu and Ma, Yining and Zhang, Jie and Gong, Yue-jiao},
  booktitle={ICLR},
  year={2024}
}

@inproceedings{ela,
  title={Exploratory landscape analysis},
  author={Mersmann, Olaf and Bischl, Bernd and Trautmann, Heike and Preuss, Mike and Weihs, Claus and Rudolph, G{\"u}nter},
  booktitle={Proceedings of the 13th annual conference on Genetic and evolutionary computation},
  year={2011}
}

@article{GP-BBOB-3,
  title={A recommender system for metaheuristic algorithms for continuous optimization based on deep recurrent neural networks},
  author={Tian, Ye and Peng, Shichen and Zhang, Xingyi and Rodemann, Tobias and Tan, Kay Chen and Jin, Yaochu},
  journal={IEEE TAI},
  year={2020},
}

@article{GP-BBOB-1,
  title={Generating new space-filling test instances for continuous black-box optimization},
  author={Mu{\~n}oz, Mario A and Smith-Miles, Kate},
  journal={Evolutionary computation},
  year={2020},
}

@inproceedings{GP-BBOB-2,
  title={Challenges of ELA-Guided Function Evolution Using Genetic Programming},
  author={Long, Fu Xing and Vermetten, Diederick and Kononova, Anna V and Kalkreuth, Roman and Yang, Kaifeng and B{\"a}ck, Thomas and van Stein, Niki},
  booktitle={IJCCI},
  year={2023}
}

@inproceedings{nlshadelbc,
  title={NL-SHADE-LBC algorithm with linear parameter adaptation bias change for CEC 2022 Numerical Optimization},
  author={Stanovov, Vladimir and Akhmedova, Shakhnaz and Semenkin, Eugene},
  booktitle={2022 IEEE CEC},
  year={2022},
}

@article{cmaes,
  title={Reducing the time complexity of the derandomized evolution strategy with covariance matrix adaptation (CMA-ES)},
  author={Hansen, Nikolaus and M{\"u}ller, Sibylle D and Koumoutsakos, Petros},
  journal={Evolutionary computation},
  year={2003},
}

@inproceedings{sepcmaes,
  title={A simple modification in CMA-ES achieving linear time and space complexity},
  author={Ros, Raymond and Hansen, Nikolaus},
  booktitle={International conference on parallel problem solving from nature},
  year={2008},
}

@article{glpso,
  title={Genetic learning particle swarm optimization},
  author={Gong, Yue-Jiao and Li, Jing-Jing and Zhou, Yicong and Li, Yun and Chung, Henry Shu-Hung and Shi, Yu-Hui and Zhang, Jun},
  journal={IEEE TC},
  year={2015},
}

@misc{uav,
      title={Benchmarking global optimization techniques for unmanned aerial vehicle path planning}, 
      author={Mhd Ali Shehadeh and Jakub Kudela},
      year={2025},
      eprint={2501.14503},
      archivePrefix={arXiv},
      primaryClass={cs.NE},
      url={https://arxiv.org/abs/2501.14503}, 
}

@article{hpob,
  title={Hpo-b: A large-scale reproducible benchmark for black-box hpo based on openml},
  author={Arango, Sebastian Pineda and Jomaa, Hadi S and Wistuba, Martin and Grabocka, Josif},
  journal={arXiv preprint arXiv:2106.06257},
  year={2021}
}

@article{proteindocking,
  title={Protein--protein docking benchmark version 4.0},
  author={Hwang, Howook and Vreven, Thom and Janin, Jo{\"e}l and Weng, Zhiping},
  journal={Proteins: Structure, Function, and Bioinformatics},
  year={2010},
}

@article{eoh,
  title={Evolution of heuristics: Towards efficient automatic algorithm design using large language model},
  author={Liu, Fei and Tong, Xialiang and Yuan, Mingxuan and Lin, Xi and Luo, Fu and Wang, Zhenkun and Lu, Zhichao and Zhang, Qingfu},
  journal={arXiv preprint arXiv:2401.02051},
  year={2024}
}

@article{Surr-RLDE,
  title={Surrogate learning in meta-black-box optimization: A preliminary study},
  author={Ma, Zeyuan and Huang, Zhiyang and Chen, Jiacheng and Cao, Zhiguang and Gong, Yue-Jiao},
  journal={arXiv preprint arXiv:2503.18060},
  year={2025}
}

@article{RLDE-AFL,
  title={Reinforcement learning-based self-adaptive differential evolution through automated landscape feature learning},
  author={Guo, Hongshu and Ma, Sijie and Huang, Zechuan and Hu, Yuzhi and Ma, Zeyuan and Zhang, Xinglin and Gong, Yue-Jiao},
  journal={arXiv preprint arXiv:2503.18061},
  year={2025}
}

@article{ALDes,
  title={Automated metaheuristic algorithm design with autoregressive learning},
  author={Zhao, Qi and Liu, Tengfei and Yan, Bai and Duan, Qiqi and Yang, Jian and Shi, Yuhui},
  journal={IEEE TEVC},
  year={2024},
}

@inproceedings{q-mamba,
title={Meta-Black-Box-Optimization through Offline Q-function Learning},
author={Zeyuan Ma and Zhiguang Cao and Zhou Jiang and Hongshu Guo and Yue-Jiao Gong},
booktitle={ICML},
year={2025},
}

@article{GLHF,
  title={Pretrained optimization model for zero-shot black box optimization},
  author={Li, Xiaobin and Wu, Kai and Zhang, Xiaoyu and Wang, Handing and Liu, Jing and others},
  journal={Advances in Neural Information Processing Systems},
  year={2024}
}

@article{metabbo-icl-5,
  title={Llamea: A large language model evolutionary algorithm for automatically generating metaheuristics},
  author={van Stein, Niki and B{\"a}ck, Thomas},
  journal={IEEE TEVC},
  year={2024},
  publisher={IEEE}
}

@inproceedings{optuna,
  title={Optuna: A next-generation hyperparameter optimization framework},
  author={Akiba, Takuya and Sano, Shotaro and Yanase, Toshihiko and Ohta, Takeru and Koyama, Masanori},
  booktitle={Proceedings of the 25th ACM SIGKDD international conference on knowledge discovery \& data mining},
  pages={2623--2631},
  year={2019}
}

@inproceedings{vizier,
  title={Google vizier: A service for black-box optimization},
  author={Golovin, Daniel and Solnik, Benjamin and Moitra, Subhodeep and Kochanski, Greg and Karro, John and Sculley, David},
  booktitle={Proceedings of the 23rd ACM SIGKDD international conference on knowledge discovery and data mining},
  pages={1487--1495},
  year={2017}
}

@article{bo,
  title={Practical bayesian optimization of machine learning algorithms},
  author={Snoek, Jasper and Larochelle, Hugo and Adams, Ryan P},
  journal={Advances in neural information processing systems},
  volume={25},
  year={2012}
}

@article{newton,
  title={A stochastic quasi-Newton method for large-scale optimization},
  author={Byrd, Richard H and Hansen, Samantha L and Nocedal, Jorge and Singer, Yoram},
  journal={SIAM Journal on Optimization},
  volume={26},
  number={2},
  pages={1008--1031},
  year={2016},
  publisher={SIAM}
}

@article{cec-moo,
  title={Multiobjective optimization test instances for the CEC 2009 special session and competition},
  author={Zhang, Qingfu and Zhou, Aimin and Zhao, Shizheng and Suganthan, Ponnuthurai Nagaratnam and Liu, Wudong and Tiwari, Santosh and others},
  year={2008},
  //publisher={Colchester, UK}
}

@inproceedings{cec2015,
  title={Problem definitions and evaluation criteria for the CEC 2015 competition on learning-based real-parameter single objective optimization},
  author={Liang, JJ and Qu, BY and Suganthan, PN and Chen, Q},
  booktitle={Proceedings of the 2015 IEEE Congress on Evolutionary Computation},
  year={2015}
}

@article{cec2017-constraint,
  title={Problem definitions and evaluation criteria for the CEC 2017 competition on constrained real-parameter optimization},
  author={Wu, Guohua and Mallipeddi, Rammohan and Suganthan, Ponnuthurai Nagaratnam},
  journal={National University of Defense Technology, Changsha, Hunan, PR China and Kyungpook National University, Daegu, South Korea and Nanyang Technological University, Singapore, Technical Report},
  volume={9},
  pages={2017},
  year={2017},
  publisher={National University of Defense Technology Changsha, China}
}

@article{chip,
  title={BBOPlace-Bench: Benchmarking Black-Box Optimization for Chip Placement},
  author={Xue, Ke and Chen, Ruo-Tong and Tan, Rong-Xi and Lin, Xi and Shi, Yunqi and Xu, Siyuan and Yuan, Mingxuan and Qian, Chao},
  journal={arXiv preprint arXiv:2510.23472},
  year={2025}
}

@article{evoprompt,
  title={Connecting large language models with evolutionary algorithms yields powerful prompt optimizers},
  author={Guo, Qingyan and Wang, Rui and Guo, Junliang and Li, Bei and Song, Kaitao and Tan, Xu and Liu, Guoqing and Bian, Jiang and Yang, Yujiu},
  journal={arXiv preprint arXiv:2309.08532},
  year={2023}
}

@article{llamea,
  title={Llamea: A large language model evolutionary algorithm for automatically generating metaheuristics},
  author={van Stein, Niki and B{\"a}ck, Thomas},
  journal={IEEE Transactions on Evolutionary Computation},
  year={2024},
  publisher={IEEE}
}

@article{alphaevolve,
  title={AlphaEvolve: A coding agent for scientific and algorithmic discovery},
  author={Novikov, Alexander and V{\~u}, Ng{\^a}n and Eisenberger, Marvin and Dupont, Emilien and Huang, Po-Sen and Wagner, Adam Zsolt and Shirobokov, Sergey and Kozlovskii, Borislav and Ruiz, Francisco JR and Mehrabian, Abbas and others},
  journal={arXiv preprint arXiv:2506.13131},
  year={2025}
}

@article{mto-benchmark,
  title={Evolutionary multitasking for single-objective continuous optimization: Benchmark problems, performance metric, and baseline results},
  author={Da, Bingshui and Ong, Yew-Soon and Feng, Liang and Qin, A Kai and Gupta, Abhishek and Zhu, Zexuan and Ting, Chuan-Kang and Tang, Ke and Yao, Xin},
  journal={arXiv preprint arXiv:1706.03470},
  year={2017}
}

@inproceedings{dynamic-benchmark,
  title={Sddobench: A benchmark for streaming data-driven optimization with concept drift},
  author={Zhong, Yuanting and Wang, Xincan and Sun, Yuhong and Gong, Yue-Jiao},
  booktitle={Proceedings of the Genetic and Evolutionary Computation Conference},
  pages={59--67},
  year={2024}
}

@article{platemo,
  title={PlatEMO: A MATLAB platform for evolutionary multi-objective optimization [educational forum]},
  author={Tian, Ye and Cheng, Ran and Zhang, Xingyi and Jin, Yaochu},
  journal={IEEE Computational Intelligence Magazine},
  volume={12},
  number={4},
  pages={73--87},
  year={2017},
  publisher={IEEE}
}

@article{funsearch,
  title={Mathematical discoveries from program search with large language models},
  author={Romera-Paredes, Bernardino and Barekatain, Mohammadamin and Novikov, Alexander and Balog, Matej and Kumar, M Pawan and Dupont, Emilien and Ruiz, Francisco JR and Ellenberg, Jordan S and Wang, Pengming and Fawzi, Omar and others},
  journal={Nature},
  volume={625},
  number={7995},
  pages={468--475},
  year={2024},
  publisher={Nature Publishing Group UK London}
}

@article{icl,
  title={Rethinking the role of demonstrations: What makes in-context learning work?},
  author={Min, Sewon and Lyu, Xinxi and Holtzman, Ari and Artetxe, Mikel and Lewis, Mike and Hajishirzi, Hannaneh and Zettlemoyer, Luke},
  journal={arXiv preprint arXiv:2202.12837},
  year={2022}
}

@article{evoprompting,
  title={Evoprompting: Language models for code-level neural architecture search},
  author={Chen, Angelica and Dohan, David and So, David},
  journal={Advances in neural information processing systems},
  volume={36},
  pages={7787--7817},
  year={2023}
}

@inproceedings{llm-moo,
  title={Large language model for multiobjective evolutionary optimization},
  author={Liu, Fei and Lin, Xi and Yao, Shunyu and Wang, Zhenkun and Tong, Xialiang and Yuan, Mingxuan and Zhang, Qingfu},
  booktitle={International Conference on Evolutionary Multi-Criterion Optimization},
  pages={178--191},
  year={2025},
  organization={Springer}
}

@inproceedings{eoh-moo,
  title={Multi-objective evolution of heuristic using large language model},
  author={Yao, Shunyu and Liu, Fei and Lin, Xi and Lu, Zhichao and Wang, Zhenkun and Zhang, Qingfu},
  booktitle={Proceedings of the AAAI Conference on Artificial Intelligence},
  volume={39},
  pages={27144--27152},
  year={2025}
}

@article{eureka,
  title={Eureka: Human-level reward design via coding large language models},
  author={Ma, Yecheng Jason and Liang, William and Wang, Guanzhi and Huang, De-An and Bastani, Osbert and Jayaraman, Dinesh and Zhu, Yuke and Fan, Linxi and Anandkumar, Anima},
  journal={arXiv preprint arXiv:2310.12931},
  year={2023}
}

@article{reevo,
  title={Reevo: Large language models as hyper-heuristics with reflective evolution},
  author={Ye, Haoran and Wang, Jiarui and Cao, Zhiguang and Berto, Federico and Hua, Chuanbo and Kim, Haeyeon and Park, Jinkyoo and Song, Guojie},
  journal={Advances in neural information processing systems},
  volume={37},
  pages={43571--43608},
  year={2024}
}

@article{evox,
  title={EvoX: A distributed GPU-accelerated framework for scalable evolutionary computation},
  author={Huang, Beichen and Cheng, Ran and Li, Zhuozhao and Jin, Yaochu and Tan, Kay Chen},
  journal={IEEE Transactions on Evolutionary Computation},
  year={2024},
  publisher={IEEE}
}

@article{pypop7,
  title={Pypop7: A pure-python library for population-based black-box optimization},
  author={Duan, Qiqi and Zhou, Guochen and Shao, Chang and Wang, Zhuowei and Feng, Mingyang and Huang, Yuwei and Tan, Yajing and Yang, Yijun and Zhao, Qi and Shi, Yuhui},
  journal={Journal of Machine Learning Research},
  year={2024}
}

@article{lsre,
  title={Instance generation for meta-black-box optimization through latent space reverse engineering},
  author={Wang, Chen and Gong, Yue-Jiao and Cao, Zhiguang and Ma, Zeyuan},
  journal={arXiv preprint arXiv:2509.15810},
  year={2025}
}

@article{ela-diversity-analysis,
  title={Exploratory landscape analysis of continuous space optimization problems using information content},
  author={Mu{\~n}oz, Mario A and Kirley, Michael and Halgamuge, Saman K},
  journal={IEEE transactions on evolutionary computation},
  volume={19},
  number={1},
  pages={74--87},
  year={2014},
  publisher={IEEE}
}

@inproceedings{algorithm-differential,
  title={Comparing algorithm selection approaches on black-box optimization problems},
  author={Kostovska, Ana and Jankovic, Anja and Vermetten, Diederick and D{\v{z}}eroski, Sa{\v{s}}o and Eftimov, Tome and Doerr, Carola},
  booktitle={Proceedings of the Companion Conference on Genetic and Evolutionary Computation},
  pages={495--498},
  year={2023}
}

@incollection{moo-paper-2,
  title={Multi-objective optimization},
  author={Deb, Kalyanmoy and Sindhya, Karthik and Hakanen, Jussi},
  booktitle={Decision sciences},
  pages={161--200},
  year={2016},
  publisher={CRC Press}
}

@article{benchofbench,
  title={Benchmark\^{} 2: Systematic Evaluation of LLM Benchmarks},
  author={Qian, Qi and Huang, Chengsong and Xu, Jingwen and Lv, Changze and Wu, Muling and Liu, Wenhao and Wang, Xiaohua and Wang, Zhenghua and Huang, Zisu and Tian, Muzhao and others},
  journal={arXiv preprint arXiv:2601.03986},
  year={2026}
}

@article{agentsci,
  title={From ai for science to agentic science: A survey on autonomous scientific discovery},
  author={Wei, Jiaqi and Yang, Yuejin and Zhang, Xiang and Chen, Yuhan and Zhuang, Xiang and Gao, Zhangyang and Zhou, Dongzhan and Wang, Guangshuai and Gao, Zhiqiang and Cao, Juntai and others},
  journal={arXiv preprint arXiv:2508.14111},
  year={2025}
}

@article{automal-bench,
  title={Amlb: an automl benchmark},
  author={Gijsbers, Pieter and Bueno, Marcos LP and Coors, Stefan and LeDell, Erin and Poirier, S{\'e}bastien and Thomas, Janek and Bischl, Bernd and Vanschoren, Joaquin},
  journal={Journal of Machine Learning Research},
  volume={25},
  number={101},
  pages={1--65},
  year={2024}
}

@article{moea/d,
  title={MOEA/D: A multiobjective evolutionary algorithm based on decomposition},
  author={Zhang, Qingfu and Li, Hui},
  journal={IEEE Transactions on evolutionary computation},
  volume={11},
  number={6},
  pages={712--731},
  year={2007},
  publisher={IEEE}
}

@article{landscape-1,
  title={PORTAL: Controllable Landscape Generator for Continuous Optimization-Part I: Framework},
  author={Yazdani, Danial and Peng, Mai and Yazdani, Delaram and Yazdi, Shima F and Omidvar, Mohammad Nabi and Sun, Yuan and Nguyen, Trung Thanh and Li, Changhe and Li, Xiaodong},
  journal={arXiv preprint arXiv:2512.00288},
  year={2025}
}

@article{landscape-2,
  title={A survey of features used for representing black-box single-objective continuous optimization},
  author={Cenikj, Gjorgjina and Nikolikj, Ana and Petelin, Ga{\v{s}}per and van Stein, Niki and Doerr, Carola and Eftimov, Tome},
  journal={Swarm and Evolutionary Computation},
  volume={101},
  pages={102288},
  year={2026},
  publisher={Elsevier}
}

@article{deep-ela,
  title={Deep-ela: Deep exploratory landscape analysis with self-supervised pretrained transformers for single-and multi-objective continuous optimization problems},
  author={Seiler, Moritz Vinzent and Kerschke, Pascal and Trautmann, Heike},
  journal={Evolutionary Computation},
  pages={1--27},
  year={2025},
  publisher={MIT Press 255 Main Street, 9th Floor, Cambridge, Massachusetts 02142, USA~…}
}

@article{deepseek-v3.2,
  title={Deepseek-v3. 2: Pushing the frontier of open large language models},
  author={Liu, Aixin and Mei, Aoxue and Lin, Bangcai and Xue, Bing and Wang, Bingxuan and Xu, Bingzheng and Wu, Bochao and Zhang, Bowei and Lin, Chaofan and Dong, Chen and others},
  journal={arXiv preprint arXiv:2512.02556},
  year={2025}
}

@article{metabbo-survey-yx,
  title={Meta-Black-Box optimization for evolutionary algorithms: Review and perspective},
  author={Yang, Xu and Wang, Rui and Li, Kaiwen and Ishibuchi, Hisao},
  journal={Swarm and Evolutionary Computation},
  volume={93},
  pages={101838},
  year={2025},
  publisher={Elsevier}
}

@article{rl-bbo-survey,
  title={Bridging evolutionary algorithms and reinforcement learning: A comprehensive survey on hybrid algorithms},
  author={Li, Pengyi and Hao, Jianye and Tang, Hongyao and Fu, Xian and Zhen, Yan and Tang, Ke},
  journal={IEEE Transactions on evolutionary computation},
  year={2024},
  publisher={IEEE}
}

@article{ml-bbo-survey,
  title={Machine Learning Algorithms for Improving Black Box Optimization Solvers},
  author={Kimiaei, Morteza and Kungurtsev, Vyacheslav},
  journal={arXiv preprint arXiv:2509.25592},
  year={2025}
}

@article{tianye-metabbo,
  title={A universal framework for automatically generating single-and multi-objective evolutionary algorithms},
  author={Tian, Ye and Qi, Xuhong and Yang, Shangshang and He, Cheng and Tan, Kay Chen and Jin, Yaochu and Zhang, Xingyi},
  journal={IEEE Transactions on Evolutionary Computation},
  year={2025},
  publisher={IEEE}
}

@article{chengran-metabbo,
  title={Learning to Evolve with Convergence Guarantee via Neural Unrolling},
  author={Gao, Jiaxin and Liu, Yaohua and Cheng, Ran and Tan, Kay Chen},
  journal={arXiv preprint arXiv:2512.11453},
  year={2025}
}

@article{du-metabbo,
  title={Meta-Black-Box Optimization with Bi-Space Landscape Analysis and Dual-Control Mechanism for SAEA},
  author={Du, Yukun and Yu, Haiyue and Xie, Xiaotong and Zheng, Yan and Zhan, Lixin and Du, Yudong and Hu, Chongshuang and Wang, Boxuan and Jiang, Jiang},
  journal={arXiv preprint arXiv:2511.15551},
  year={2025}
}

@inproceedings{wukai-metabbo,
  title={Enhancing Zero-Shot Black-Box Optimization via Pretrained Models with Efficient Population Modeling, Interaction, and Stable Gradient Approximation},
  author={Han, Muqi and Li, Xiaobin and Wu, Kai and Zhang, Xiaoyu and Wang, Handing},
  booktitle={The Thirty-ninth Annual Conference on Neural Information Processing Systems},
  year={2025}
}

@inproceedings{
xueke-metabbo,
title={Sequential Multi-Agent Dynamic Algorithm Configuration},
author={Chen Lu and Ke Xue and Lei Yuan and Yao Wang and Yaoyuan Wang and Fu Sheng and Chao Qian},
booktitle={The Thirty-ninth Annual Conference on Neural Information Processing Systems},
year={2025},
}

@article{pei2025libog,
  title={LiBOG: Lifelong Learning for Black-Box Optimizer Generation},
  author={Pei, Jiyuan and Mei, Yi and Liu, Jialin and Zhang, Mengjie},
  journal={arXiv preprint arXiv:2505.13025},
  year={2025}
}

@article{wu2025learning,
  title={Learning to transfer for evolutionary multitasking},
  author={Wu, Sheng-Hao and Huang, Yuxiao and Wu, Xingyu and Feng, Liang and Zhan, Zhi-Hui and Tan, Kay Chen},
  journal={IEEE Transactions on Cybernetics},
  year={2025},
  publisher={IEEE}
}

@inproceedings{
metaes,
title={Discovering Evolution Strategies via Meta-Black-Box Optimization},
author={Robert Tjarko Lange and Tom Schaul and Yutian Chen and Tom Zahavy and Valentin Dalibard and Chris Lu and Satinder Singh and Sebastian Flennerhag},
booktitle={The Eleventh International Conference on Learning Representations },
year={2023},

}

@inproceedings{sdms-pso,
  title={A self-adaptive dynamic particle swarm optimizer},
  author={Liang, Jing J and Guo, L and Liu, R and Qu, Bo-Yang},
  booktitle={2015 IEEE Congress on Evolutionary Computation (CEC)},
  pages={3206--3213},
  year={2015},
  organization={IEEE}
}

@article{sahl-pso,
  title={Self-Adaptive two roles hybrid learning strategies-based particle swarm optimization},
  author={Tao, Xinmin and Li, Xiangke and Chen, Wei and Liang, Tian and Li, Yetong and Guo, Jie and Qi, Lin},
  journal={Information Sciences},
  volume={578},
  pages={457--481},
  year={2021},
  publisher={Elsevier}
}
\bibliographystyle{icml2026}

\newpage
\appendix
\onecolumn
\section{Prompts in EoB}\label{appx:prompts}

\begin{figure}
    \centering
    \includegraphics[width=\linewidth]{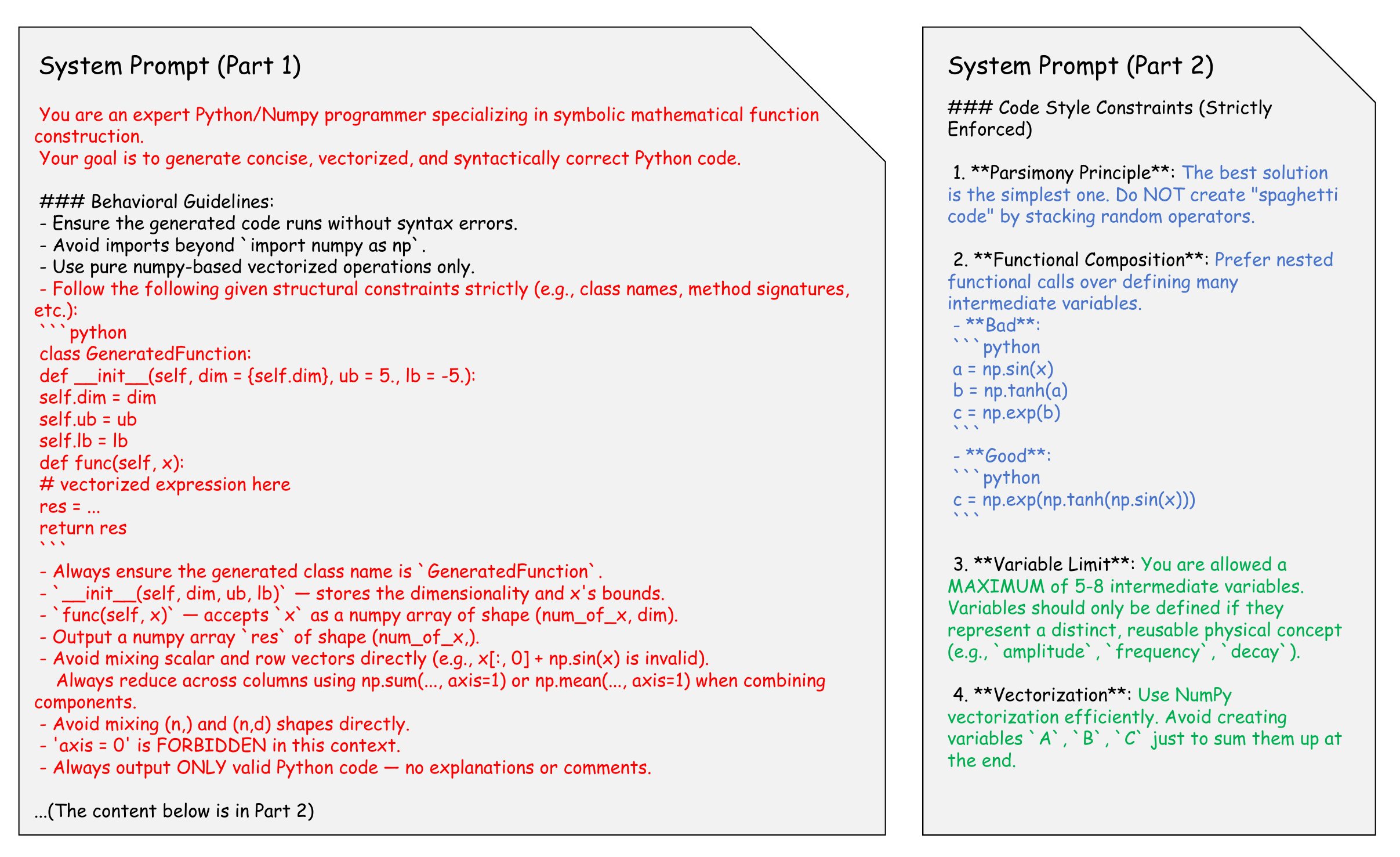}
    \caption{Code Generation Guideline~(System Prompt)}
    \label{fig:A.1-sys-prompt}
\end{figure}

\subsection{Code Generation Guideline~(System Prompt)}\label{appx:Guideline}
This prompt serves as the foundational rule set for the LLM, strictly regulating the format, interface, and stylistic quality of the generated Python code. The detailed design of the prompts is shown in fig~\ref{fig:A.1-sys-prompt}. Each part of the prompt and its description will be highlighted in three different colors, as described below: 

\begin{itemize}
    \item \textcolor{red}{Interface Standardization:} To ensure the generation of executable and mathematically coherent benchmarks, this prompt conditions the LLM as a specialized NumPy expert, enforcing a strict schema where outputs must be encapsulated within a `GeneratedFunction` class supporting vectorized inputs. Crucially, the prompt imposes rigorous syntactic and stylistic regularization to guarantee solvability and readability: it mandates shape consistency (e.g., prohibiting reduction along the batch dimension to preserve sample independence).
    \item \textcolor[RGB]{0,112,192}{Parsimony Principle: } To prevent the generated function code from becoming "spaghetti-like," our prompt design requires the LLM to avoid defining a large number of intermediate variables when generating nested structured functions. Instead, it should construct them by directly combining functions.
    \item \textcolor[RGB]{0,176,80}{Concise, Efficient, and Meaningful: } We restrict the number of intermediate variables generated by the LLM during the code generation process, requiring each intermediate variable to correspond to a specific physical concept to ensure meaningfulness. Additionally, we mandate the efficient use of NumPy operations during code generation to improve performance.
\end{itemize}

\subsection{Initialization}\label{appx:init_prompt}
\begin{figure}
    \centering
    \includegraphics[width=\linewidth]{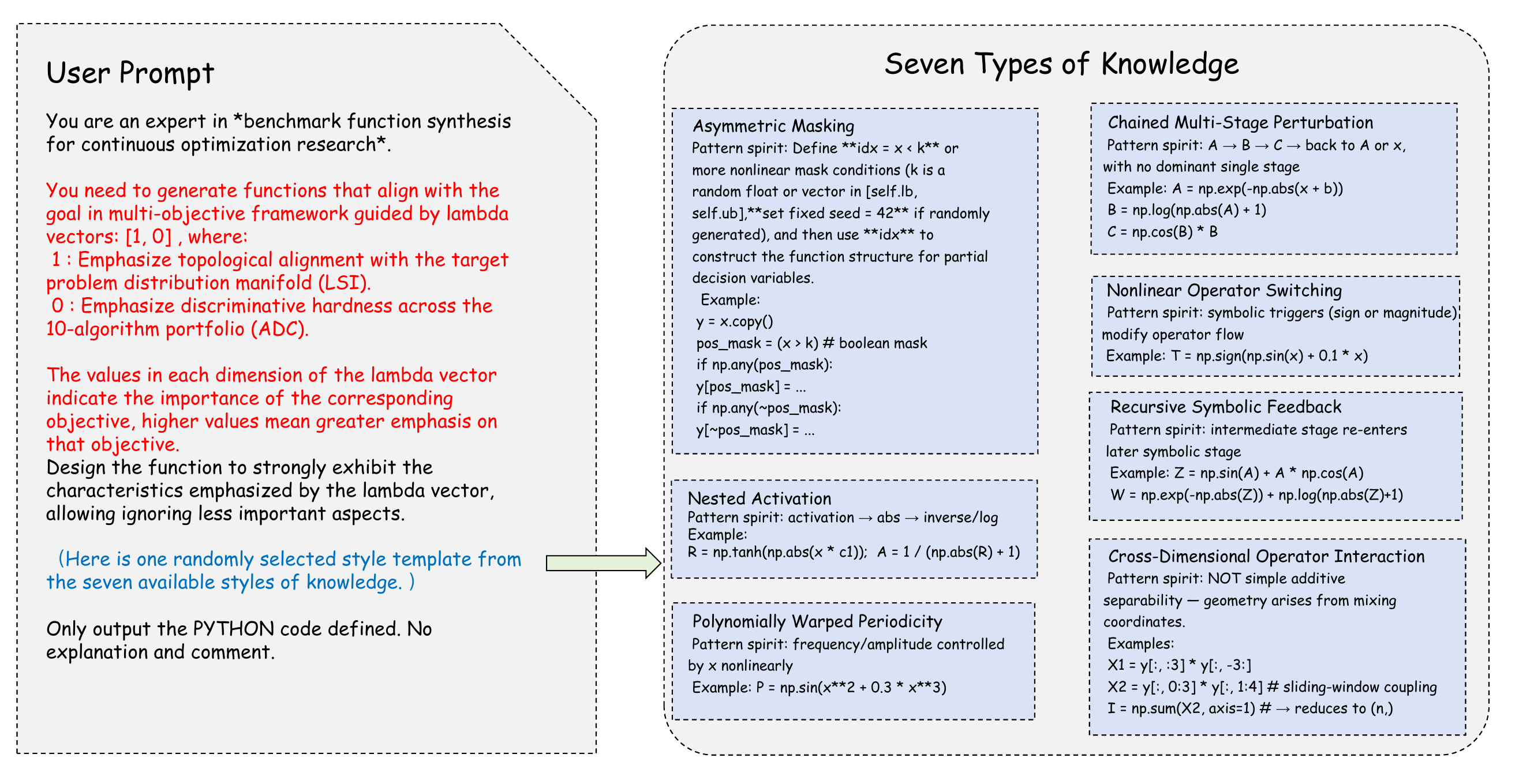}
    \caption{Initialization Prompt~(User)}
    \label{fig:A.2-usr-prompt}
\end{figure}

In the strategy described in Section~\ref{sec:3.2.1}, we employ a carefully constructed \emph{Init\_Prompt} to help LLM grasps the essential context information, such as its role (as an optimization expert and benchmark designer) and its specific task (generating program code for the objective function). The knowledge regarding the code generation of the objective function in the \emph{Init\_Prompt} is provided by the \emph{System\_Prompt} in Appendix~\ref{appx:Guideline}, while the knowledge related to the roles of the LLM as a benchmark designer and optimization expert is provided by the \emph{User\_Prompt}. The content of the \emph{User\_Prompt} is shown in Fig~\ref{fig:A.2-usr-prompt}. 

The \emph{User\_Prompt} consists of two major parts, which are highlighted in different colors in the following content and in Fig~\ref{fig:A.2-usr-prompt} : 
\begin{itemize}
    \item \textcolor[RGB]{255,0,0}{Generation Preference.} This part drives the generation of the initial individual by explicitly defining the multi-objective optimization context. The prompt explicitly maps the reference vector $\vec{\lambda_i} = [\lambda_{i,1}, \lambda_{i,2}]$ dimensions to semantic objectives: $\lambda_{i,1}$: Emphasizes topological alignment with the target problem distribution~(LSI); $\lambda_{i,2}$: Emphasizes discriminative hardness, encouraging the creation of "strategy-breaker" landscapes that differentiate between algorithms~(ADC). Guided by this part, LLM can design the function that strongly exhibits the characteristics emphasized by $\vec{\lambda_i}$, while ignoring less important aspects.
    \item \textcolor[RGB]{0,112,192}{Population Diversity.} To address the issue of LLM generating a large number of highly similar function instances due to a lack of landscape knowledge, which leads to a loss of diversity in the initial population, we constructed a knowledge base comprising seven types of landscape styles (as shown in the right part of Fig~\ref{fig:A.2-usr-prompt}). By injecting random landscape style knowledge into the prompts used to generate initial individuals, we enhance both the code diversity and landscape diversity of the initial population. The seven types of landscape style knowledge are introduced as follows: 
    \begin{itemize}
        \item \textbf{Asymmetric Masking.} The Asymmetric Masking style explicitly disrupts global symmetry by applying distinct symbolic operations to sub-regions of the search space defined by boolean conditions (e.g., $x < k$). By forcing the LLM to use np.where or boolean indexing, this style creates landscapes with sharp regional shifts and discontinuities, challenging algorithms that assume global rotational invariance. 
        \item \textbf{Nested Activation.} The Nested Activation style (often implemented via rational composition) creates landscapes with steep, controlled basins. It enforces a specific operator pipeline, typically composing activation functions with rational transformations (e.g., $f(x) = \frac{1}{|\tanh(x)| + 1}$) to generate narrow valleys that test algorithmic stability without causing numerical overflow. 
        \item \textbf{Polynomially Warped Periodicity.} The Polynomially Warped Periodicity style  modulates the frequency or amplitude of periodic functions using nonlinear polynomial terms. Unlike standard Rastrigin-like functions, this style produces "chirp-like" structures (e.g., $\sin(x^2 + 0.3x^3)$) where the gradient frequency changes dynamically across the search space, challenging step-size adaptation mechanisms.
        \item \textbf{Chained Multi-stage Perturbation.} The Chained Multi-stage Perturbation style fosters feature mixing by enforcing a cyclical composition of disparate mathematical operators (e.g., exponential decay feeding into logarithmic scaling, then into trigonometric oscillation). This ensures that no single operator dominates the global trend, resulting in hybrid topologies that are difficult to classify under standard BBOB categories.
        \item \textbf{Recursive Symbolic Feedback.} The Recursive Symbolic Feedback style generates fractal-like micro-structures by allowing intermediate symbolic variables to re-enter calculations later in the pipeline (e.g., $Z = \sin(A) + A \cdot \cos(A)$). This recursion creates deep variable dependencies where small input changes propagate into high-frequency local optima.
        \item \textbf{Nonlinear Operator Switching.} The Nonlinear Operator Switching style utilizes symbolic triggers (such as magnitude or sign) to modify operator flow, creating "manifold switching" effects where the landscape topology changes smoothly but drastically.
        \item \textbf{Cross-Dimensional Operator Interaction} The Cross-dimensional Operator Interaction module mandates coordinate mixing (e.g., sliding window coupling or product interactions like $x_i \cdot x_j$ and $x[:k] * x[-k:]$). Uniquely, this style permits temporary dimensional reduction (e.g., np.sum along the feature axis) to create interaction terms, which are then broadcast back to the original shape, thereby invalidating algorithms that rely on independent coordinate search.
    \end{itemize}
\end{itemize}

\begin{figure}
    \centering
    \includegraphics[width=\linewidth]{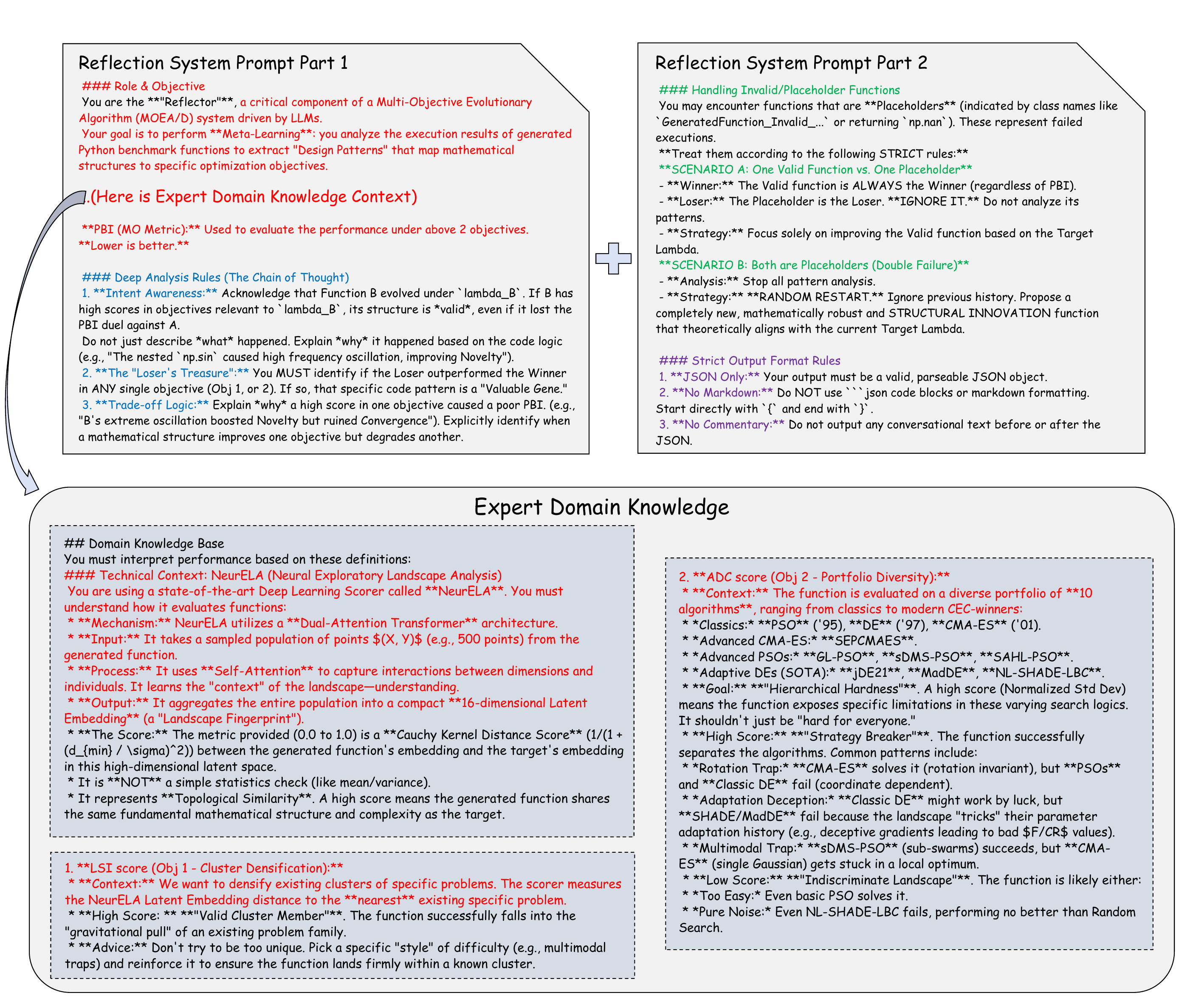}
    \caption{Reflection Stage System Prompt}
    \label{fig:A.3-sys-prompt}
\end{figure}

In conclusion, the \emph{Init\_Prompt} relies on a synergistic prompting strategy that integrates the \emph{System\_Prompt} and the \emph{User\_Prompt}. The \emph{System\_Prompt} acts as a strict Code Generation Guideline, imposing strict syntactic constraints, interface standardization, and stylistic regularization to guarantee the executability and numerical stability of the generated code. The \emph{User\_Prompt} injects the multi-objective reference vectors ($\vec{\lambda}$) and specific Landscape Construction Knowledge that randomly selected from the seven landscape styles detailed above. This dual-prompt framework effectively conditions the LLM to function as a domain expert, ensuring that the resulting initial population comprises objective function programs that are not only computationally compliant but also exhibit rich and distinct diversity, laying a robust foundation for the subsequent evolutionary process.

\begin{figure}
    \centering
    \includegraphics[width=\linewidth]{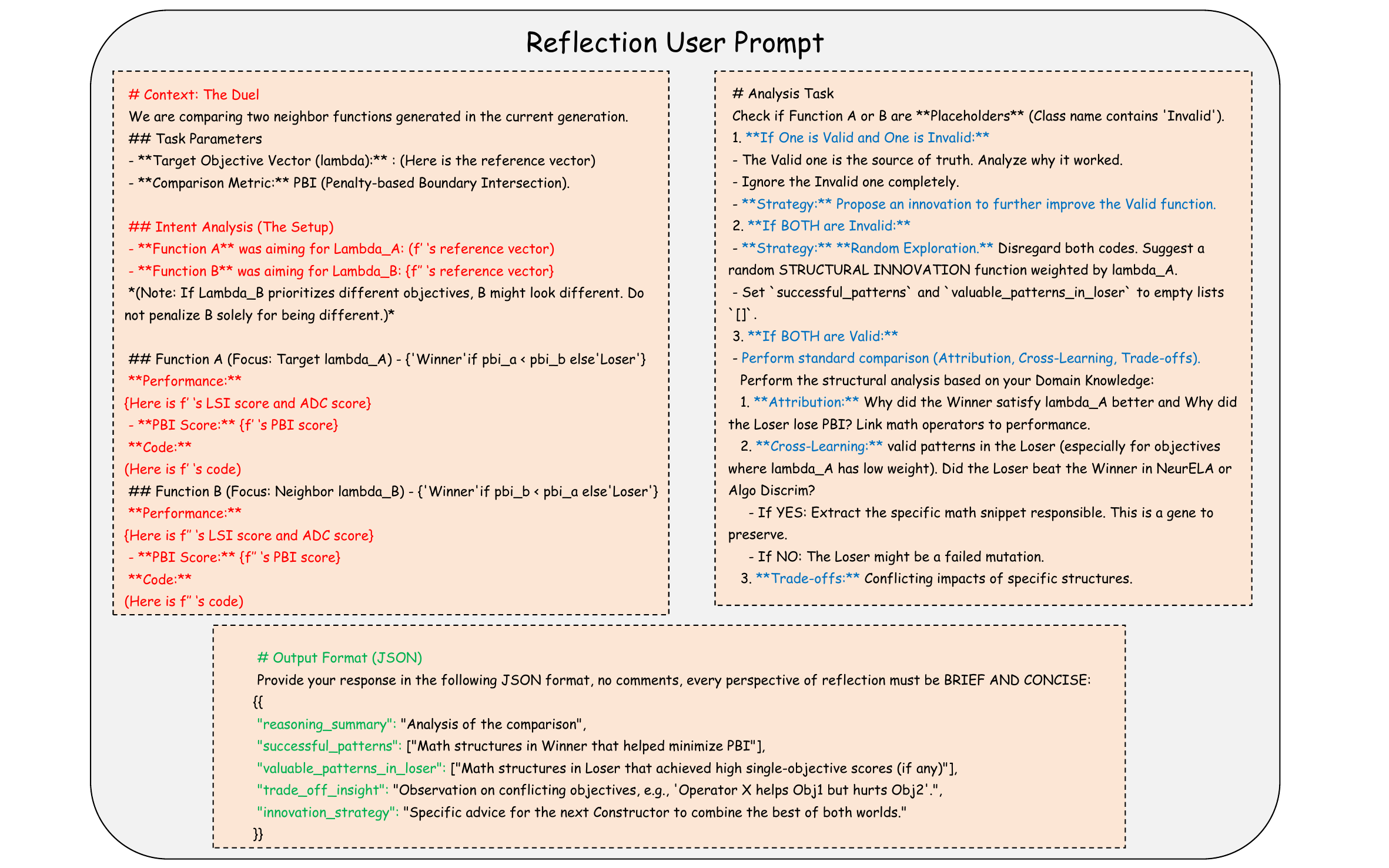}
    \caption{Reflection User Prompt}
    \label{fig:A.3-usr-prompt}
\end{figure}

\subsection{Reflection~\&~Reproduction}
As mentioned in Section~\ref{sec:3.2.3}, the reflection and reproduction stages serve as the cognitive core of our evolutionary framework, simulating the crossover and mutation mechanisms of traditional evolutionary algorithms but operating within the semantic space of code analysis. The core objective is to convert numerical multi-objective performance into actionable coding insights. 
\subsubsection{Reflection stage}\label{appx:reflection}
To enable the LLM to perform high-level reasoning and comparative analysis, we designed a two-part \emph{Reflection\_Prompt} that explicitly instructs the LLM to play the role of a senior optimization and landscape design expert. This enables LLM to interpret complex code structures based on the two objectives defined in Section~\ref{sec:3.2.2}, as well as knowledge of landscape theory and optimization principles. The \emph{Reflection\_Prompt} consists of the following two components:
\begin{itemize}
    \item \textbf{System Prompt.} The system prompt serves as the foundational rule set for the Reflection stage. It transforms raw execution program into actionable design intelligence by analyzing "Winner vs. Loser" pairs to extract causal links between mathematical structures and multi-objective metrics. The prompt consists of four streamlined modules which highlighted with different color in Fig~\ref{fig:A.3-sys-prompt} and following content :
    \begin{itemize}
        \item \textcolor{red}{Role \& Obejective~(with Domain Knowledge).} This module establishes the role of an expert "Reflector" and injects a specialized Domain Knowledge Base. It provides theoretical context for NeurELA's structure and mechanism,  LSI score and ADC score, ensuring all analysis is technically grounded.
        \item \textcolor[RGB]{0,112,192}{Deep Analysis Rules.} This module requires the LLM to conduct a three-fold in-depth reflection (when comparing functions $f'$ and $f''$, a candidate is labeled as the winner if its PBI is higher, or as the loser otherwise): 1) Successful Pattern: Identify the major code-level differences between the winner and the loser, summarize such knowledge, and relate it to landscape characteristics. 2) Gene Preservation: If the loser scores higher in either LSI or ADC, the LLM is asked to analyze the reasons behind it and preserve such valuable knowledge. 3) Bi-objective Tradeoff: Recognize and analyze the positive/negative impacts of certain parts of the program’s code on LSI and ADC, with a special focus on conflict cases—for example, adding a mathematical operation may increase LSI but harm ADC. Such cases serve as the basis for crossover operations.
        \item \textcolor[RGB]{0,176,80}{Handling Invalid Functions.} This module ensures robustness in runtime errors by defining strict branching logic. In the scenario of "one valid item and one placeholder item," it focuses solely on improving the valid function. In the scenario of "dual failure," it triggers a "random restart," instructing the model to abandon the current path and propose entirely new structural innovations. Such cases serve as the basis for mutation operations.
        \item \textcolor[RGB]{112,48,160}{Output Formatting Constraints.} To facilitate automated processing, this module enforces a rigid JSON-only output format. It strictly forbids Markdown formatting or conversational fillers, ensuring the extracted insights and improvement strategies can be directly parsed and utilized by the subsequent Reproduction stage.
    \end{itemize} 
    \item \textbf{User Prompt.} The user prompt serves as a dynamic context injector, incorporating details about the selected two candidates, $f'$ and $f''$, from the neighborhood set into a templated prompt. It dynamically constructs a task prompt for each function evolution duel. The user prompt can be divided into the following three parts, with the content of each section highlighted in different colors in the text below and in Fig~\ref{fig:A.3-usr-prompt}:
    \begin{itemize}
        \item \textcolor{red}{Contextual Instantiation (The Duel Setup).}  This module provides the raw material for analysis by injecting the source codes, LSI, ADC, and PBI Scores of the paired candidates $f'$ and $f''$. Crucially, it explicitly provides the specific reference vector ($\vec{\lambda}_f'$) and the neighbor's vector ($\vec{\lambda}_f''$). This enables the LLM to integrate the preferences of its reference vector when conducting comparative analysis of a pair of functions, thereby extracting insights and experience. This prevents the LLM from relying solely on absolute LSI or ADC scores and drawing extreme evolutionary insights.
        \item \textcolor[RGB]{0,112,192}{Scenario-Based Evolution.} This module explicitly marks the validity status of each function (identifying the "Invalid" category), thereby requiring the LLM to provide evolutionary insights in different directions. The design of this process is similar to the corresponding section in the reflection system prompt mentioned above. It instructs the LLM to perform the following operations: conduct standard comparative analysis when both functions are valid; execute "innovative improvement" guided by the reference vector when only a single valid function remains; or trigger "random exploration" in cases of dual failure.
        \item \textcolor[RGB]{0,176,80}{Strategy Output Format.} This module requires mapping the final analysis to the specified JSON output fields. The content of the JSON fields is largely consistent with the analytical rules of the reflection system prompt. It instructs the LLM to dissect the "winner" to extract successful patterns (i.e., structures that minimize the PBI), output as "successful patterns"; to mine valuable patterns from the "loser", specifically, certain code snippets that may score high on a single objective (such as high ADC) but perform poorly overall in terms of PBI, output as "valuable patterns in loser"; to analyze code snippets where potential improvement conflicts exist, output as "trade off insight"; and finally, to output the summarized insights as "innovation strategy".
    \end{itemize}
\end{itemize}
After completing the design of the aforementioned system prompt and user prompt, during the evolution process, the reflector dynamically injects into the user prompt the source code, LSI, ADC, and PBI scores required for reflection, along with the corresponding reference vector. It then combines the two prompts and calls the LLM to perform all intermediate analyses and reasoning, organizing the response content in a JSON-like format. This includes a summary of the overall reasoning process, condensed analysis results for the above case-by-case reflection, and suggestions/instructions for generating the next generation offspring program.

\subsubsection{Reproduction Stage}\label{appx:reproduction}
Based on the JSON-like reflections obtained in the reflection stage, the reproduction stage requires the LLM to refer to and follow these useful analyses or improvement suggestions to generate new program candidates. In this stage, a new role must be defined for the LLM, and the JSON-formatted reflections generated in the reflection stage need to be unpacked and injected into a prompt template that includes the reference vector and the codes of the two parent. Similarly, the prompt for this stage also consists of two parts, as illustrated in Fig~\ref{fig:A.3.2-prompt} and described below: 
\begin{figure}
    \centering
    \includegraphics[width=\linewidth]{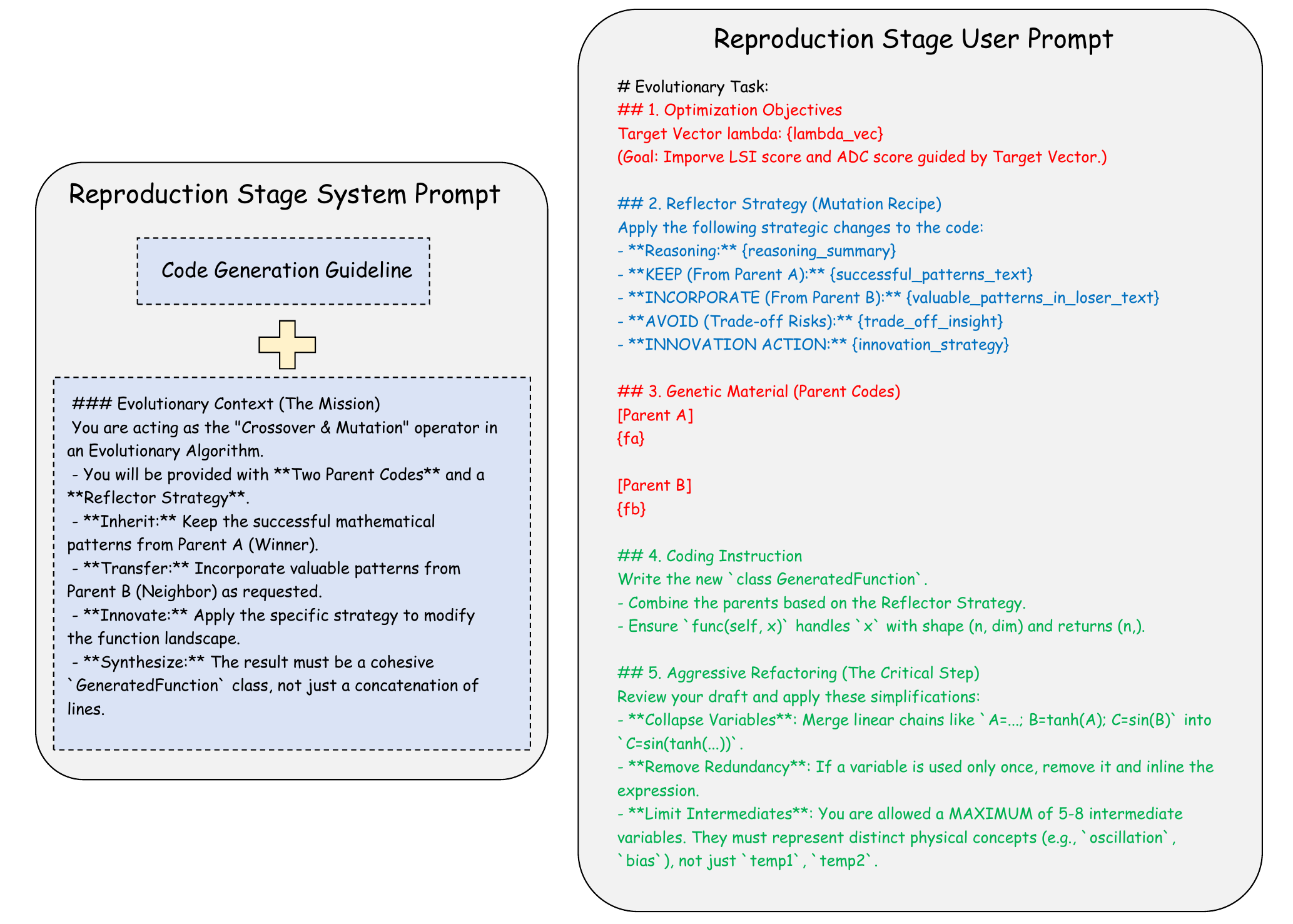}
    \caption{Reproduction Stage Prompt}
    \label{fig:A.3.2-prompt}
\end{figure}
\begin{itemize}
    \item \textbf{System Prompt.} In addition to strictly adhering to the code generation guidelines defined in Appendix~\ref{appx:Guideline}, the system prompt for the reproduction stage clearly defines its role as an evolutionary operator (crossover and mutation). This module, guided by a specific reflection strategy, enables the LLM to process two parent codes and synthesize a new offspring. The generation process follows a strict four-step protocol: inheritance (retain the robust mathematical structure of the winner), transfer (assimilate beneficial "genes" from the neighbor), innovation (apply strategic perturbations), and synthesis. Crucially, the prompt requires the final output to be a unified, mathematically coherent 'GeneratedFunction' class, explicitly prohibiting the mechanical concatenation of unrelated code segments.
    \item \textbf{User Prompt.} The user prompt for the reproduction stage is dynamically constructed by parsing and unpacking the structured JSON insights derived from the Reflection stage, while simultaneously injecting the source codes of the two parent candidates and target reference vector. This prompt is structurally organized into three distinct operational modules highlighted in different colors in Fig~\ref{fig:A.3.2-prompt} and following content: 
    \begin{itemize}
        \item \textcolor[RGB]{255,0,0}{Optimization Context and Genetic Material.} This module establishes the boundary conditions for the new generation. It explicitly defines the Optimization Objectives (including the target reference vector $\vec{\lambda}$ and specific landscape goals) and provides the raw Genetic Material by embedding the full source codes of two parent ($f_A$ and $f_B$). This ensures the LLM has direct access to both the specific directional goal and the exact syntax of the predecessor functions.
        \item \textcolor[RGB]{0,112,192}{The Reflector Strategy.} Acting as the core evolutionary directive, this module translates the insights extracted from the previous stage into actionable instructions. It presents a structured "Evolution Recipe" that explicitly dictates: 1) Keep: Which successful patterns from Parent A must be preserved. 2) Incorporate: Which specific "genes" (code snippets) from Parent B should be transferred. 3) Avoid: Which trade-off risks to mitigate. 4) Innovation: The specific landscape modification strategy to apply. This ensures that the crossover is not random, but a directed synthesis based on the Reflector's analysis. 
        \item \textcolor[RGB]{0,176,80}{Coding Instruction and Aggressive Refactoring.} This module enforces strict implementation standards to maintain code quality. The rules for code writing and refactoring in the prompt are similar to those in the code generation guidelines of the system prompt and will not be elaborated on further here.
    \end{itemize}
\end{itemize}




\end{document}